\documentclass[10pt,twocolumn,letterpaper]{article}

\usepackage[pagenumbers]{cvpr} 

\usepackage{graphicx}
\usepackage{amsmath}
\usepackage{amssymb}
\usepackage{booktabs}

\usepackage[pagebackref,breaklinks,colorlinks]{hyperref}

\usepackage[capitalize]{cleveref}
\crefname{section}{Sec.}{Secs.}
\Crefname{section}{Section}{Sections}
\Crefname{table}{Table}{Tables}
\crefname{table}{Tab.}{Tabs.}

\def\cvprPaperID{*****} 
\def\confName{CVPR}
\def\confYear{2022}

\begin{document}

\title{PP-Matting: High-Accuracy Natural Image Matting}

\author{Guowei Chen, Yi Liu, Jian Wang, Juncai Peng, Yuying Hao,\\ Lutao Chu, Shiyu Tang, Zewu Wu, Zeyu Chen, Zhiliang Yu,\\ Yuning Du, Qingqing Dang,  Xiaoguang Hu, Dianhai Yu\\
        Baidu Inc.\\
        {\tt\small \{chenguowei01, liuyi22\}@baidu.com}
}
\maketitle

\begin{abstract}
Natural image matting is a fundamental and challenging computer vision task. It has many applications in image editing and composition. Recently, deep learning-based approaches have achieved great improvements in image matting. However, most of them require a user-supplied trimap as an auxiliary input, which limits the matting applications in the real world. Although some trimap-free approaches have been proposed, the matting quality is still unsatisfactory compared to trimap-based ones. Without the trimap guidance, the matting models suffer from foreground-background ambiguity easily, and also generate blurry details in the transition area. In this work, we propose PP-Matting, a trimap-free architecture that can achieve high-accuracy natural image matting. Our method applies a high-resolution detail branch (HRDB) that extracts fine-grained details of the foreground with keeping feature resolution unchanged. Also, we propose a semantic context branch (SCB) that adopts a semantic segmentation subtask. It prevents the detail prediction from local ambiguity caused by semantic context missing. In addition, we conduct extensive experiments on two well-known benchmarks: Composition-1k and Distinctions-646. The results demonstrate the superiority of PP-Matting over previous methods. Furthermore, we provide a qualitative evaluation of our method on human matting which shows its outstanding performance in the practical application. The code and pre-trained models will be available at PaddleSeg: \href{https://github.com/PaddlePaddle/PaddleSeg}{https://github.com/PaddlePaddle/PaddleSeg}.

\end{abstract}

\begin{figure}[t]
\centering
\includegraphics[width=0.4\textwidth]{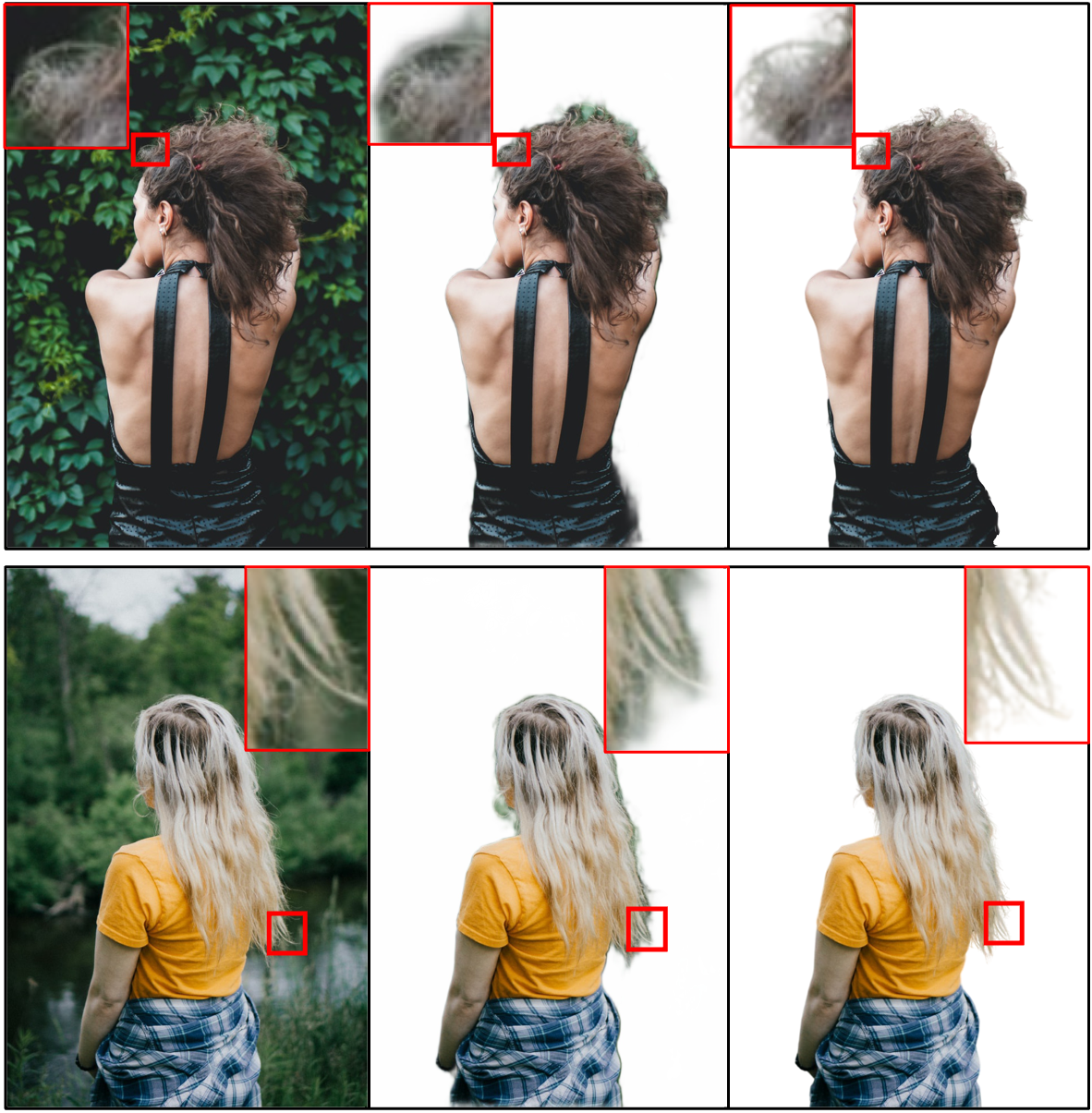}
\caption{Comparison of segmentation and matting. The original image is on the left, the segmentation results is in the middle, and the matting results is on the right.}
\label{fig:seg_mat}
\end{figure}

\section{Introduction}
\label{sec:intro}
Natural image matting refers to accurately estimating the per-pixel opacity of the target foreground from an image. As a fundamental and challenging computer task, image matting has received great interest
from both academia and industry and has been extensively studied
in the past decades~\cite{aksoy2018semantic,aksoy2017designing,bai2007geodesic,chen2013knn,chuang2001bayesian,feng2016cluster,gastal2010shared,grady2005random,he2011global,levin2007closed,levin2008spectral,ruzon2000alpha}. Intuitively, image matting generates more natural and delicate foreground than semantic segmentation~\cite{chu2022pp, liu2021paddleseg} as shown in Figure~\ref{fig:seg_mat}. It is a key technique of image/video editing and composition, which has been widely used in virtual reality, augmented reality, and film production~\cite{boda2018survey}. 

Formally, an image $I \in \mathbb{R}^{H \times W \times 3}$ is a composite of a foreground image $F \in \mathbb{R}^{H \times W \times 3}$ and a background image $B \in \mathbb{R}^{H \times W \times 3}$. The color of the i-th pixel in the given image can be formulated as a linear combination equation of foreground and background colors with alpha matte $\alpha \in \mathbb{R}^{H \times W}$,
\begin{equation}
  I^i = \alpha^i F^i + (1 - \alpha^i)B^i
  \label{eq:important}
\end{equation}
where $\alpha^i$ is the foreground opacity at pixel $i$. The image matting problem is highly ill-posed with 7 values to be solved, but only 3 values are known for each pixel in a given RGB image. To decrease
the difficulty of this problem, matting approaches usually require an auxiliary input, e.g. trimap, in addition to the original image~\cite{bai2007geodesic,chen2013knn,chuang2001bayesian,gastal2010shared,grady2005random,he2011global,levin2007closed}. The trimap is a rough segmentation of the image into three parts: foreground, background and transition (regions with unknown opacity). Although deep learning-based approaches have been proven powerful in the field of image matting, most early works still need to take the extra trimap as the input, which limits the application of the network. Most users have difficulty in creating a trimap. In some cases, it is not feasible to provide the trimap, e.g. live video.

Recently, trimap-free approaches have emerged~\cite{zhang2019late,yu2020high,chen2018semantic,deora2021salient,liu2020boosting,yu2021cascade,ke2020green,li2021deep,li2022bridging, lin2021real, sengupta2020background}, which do not require a user-supplied trimap as the auxiliary input. Lin et al. \cite{lin2021real} and Sengupta et al. \cite{sengupta2020background} replace the trimap with a background image that is easier to obtain. However, if the provided background image is much different from the real background, the performance deteriorates dramatically, e.g. in a dynamic environment. Thus, using the background image as input is still not flexible, and using a single image as input 
would be promising in practice. The approaches in \cite{sharma2020alphanet, deora2021salient, chen2018semantic} separate the trimap-free prediction into a semantic segmentation task and a matting task. The segmentation task takes the single image to generate a 3-classes mask that is taken as a trimap automatically, and then it is concatenated with the original image as input to the matting task. However, the multi-stage approaches could introduce the problem of accumulative error, if the former segmentation task outputs an incorrect mask. Besides, the training of the multi-stage model is difficult.

To overcome the multi-stage limitation, single-stage approaches have been proposed~\cite{qiao2020attention,li2022bridging, li2021deep}, where the single stage means they can be trained in an end-to-end way without the intermediate result generation. Qiao et al.~\cite{qiao2020attention} propose spatial attention on appearance cues to filtrate image texture details and implicit global guidance, which is challenging to generate alpha well. Li et al.~\cite{li2022bridging, li2021deep} design a glance branch and a focus branch to make semantic prediction and detail prediction, but the two branches do not have enough interaction. The downsample-upsample design of the focus branch also makes the details blurry in high-resolution images. Therefore, the previous approaches have difficulty in aggregating semantic and detail information, so they fail to carry out high-accuracy image matting.

In this work, we propose a semantic-aware architecture named PP-Matting, which achieves high-accuracy image matting. To obtain fine-grained details of the foreground, our method applies a high-resolution detail branch (HRDB). It keeps high resolution while extracting features in different levels rather than a downsample-upsample encoder-decoder structure. Due to a lack of semantics, the detail prediction is prone to foreground-background ambiguity. Thus, we propose a semantic context branch (SCB) that employs the segmentation subtask to ensure the semantic correctness of details. Then, the final alpha matte is obtained by fusing the detail prediction from HRDB with the semantic map of SCB. Furthermore, the extensive experiments on Composition-1k and Distinctions-646 demonstrate our state-of-the-art performance compared to previous methods.

Our main contributions are summarized as follows:
\begin{itemize}
  \item We propose PP-Matting, a high-accuracy matting network, which takes a single image as input without any auxiliary information. The whole network can be trained easily in an end-to-end way.
  \item We propose the two-branch architecture that extracts detail and semantic features efficiently in parallel. With a guidance flow mechanism, the proper interaction of the two branches helps the network achieve better semantic-aware detail prediction.
  \item We evaluate PP-Matting on Composition-1k and Distinctions-646 datasets. The results demonstrate the superiority of PP-Matting over other methods. Another experiment on human matting also shows its outstanding performance in practical application.
\end{itemize}

\section{Related Work}
Natural image matting is an active topic in both academia and industry. In general, there are roughly two categories of matting approaches: traditional matting and deep-learning matting.

\textbf{Traditional matting.} Traditional approaches usually take an RGB-channel image together with additional inputs, i.e. trimap or scribbles. The trimap is a rough segmentation of the image into three parts: foreground, background, and transition region, while the scribbles indicate a small number of pixels belonging to the foreground or background. The purpose of the additional inputs is to reduce the difficulty of alpha mattes estimation, which is a highly ill-posed problem. According to how the additional inputs are used, traditional matting approaches are further divided into two categories: sampling-based approaches and affinity-based approaches. 
Sampling-based approaches \cite{chuang2001bayesian,feng2016cluster,gastal2010shared,he2011global,ruzon2000alpha} infer the alpha values of the transition region by a class model which is built by using the color features with additional low-level features of the sampled pixels. The accuracy of these methods often depends on the quality of the trimap. Affinity-based approaches \cite{aksoy2017designing,aksoy2018semantic,bai2007geodesic,chen2013knn,grady2005random,levin2007closed,levin2008spectral} leverage pixel similarities calculated by the spatial and color features to propagate the alpha values of the known foreground and background pixels to the transition regions. Due to the spatial proximity, affinity-based approaches can generate a more smooth matte than sampling-based ones.

\textbf{Deep-learning matting.} Over the past decade, deep-learning-based approaches have become dominant in the matting field. In early works, deep learning is an intermediate step in the whole matting process. Shen et al.~\cite{shen2016deep} use a deep learning model to predict the trimap and then estimate alpha matting through a traditional matting approach. Cho et al.~\cite{cho2016natural} take the results of traditional matting and normalized RGB colors as inputs into a CNN model to predict refined alpha mattes. Xu et al.~\cite{xu2017deep} adopt a VGG-like model to directly predict the alpha matte by using an RGB image and a trimap, which is regarded as a pioneer work in deep learning matting. After that, many trimap-based matting approaches using deep learning are proposed. Tang et al. \cite{tang2019learning} propose a hybrid sampling-based and learning-based matting approach. Hou et al. \cite{hou2019context} present a context-aware natural image matting method for simultaneous foreground and alpha mattes estimation. Lu et al. \cite{lu2019indices} propose a flexible network module named IndexNet, which dynamically generates indices conditioned on the feature map. Cai et al. \cite{cai2019disentangled} propose AdaMatting, which disentangles the matting task into trimap adaptation and alpha estimation. Li et al. \cite{li2020natural} use guided contextual attention to propagate high-level opacity information globally based on the learned low-level affinity. Sun et al. \cite{sun2021semantic} learn 20 semantic classes of matting patterns to improve matting. Yu et al. \cite{yu2020high} propose HDMatt, which processes high-resolution inputs in a patch-based crop-and-stitch manner. However, all of these approaches are based on trimap, which is difficult to be provided by inexperienced users so that the applications are limited.

To eliminate the trimap, Lin et al. \cite{lin2021real} and Sengupta et al. \cite{sengupta2020background} use the background image as an auxiliary input. However, if the provided background image is much different from the real background, the performance deteriorates dramatically. Besides, such an auxiliary input also limits the application in practice. Zhang et al. \cite{zhang2019late} use two decoder branches for the foreground
and background classification, respectively, which provides more degrees of freedom than a single decoder branch for the network to obtain better
alpha values during training. Yu et al. \cite{qiao2020attention} employ spatial and channel-wise attention to integrate appearance cues and pyramidal features. Chen et al. \cite{chen2018semantic} and Deora et al. \cite{deora2021salient} use a semantic network to predict trimap and then concatenate it with RGB images as the input of the matting network. Liu et al. \cite{liu2020boosting}, Yu et al. \cite{yu2021cascade}, and Ke et al. \cite{ke2020green} get the alpha mattes in a coarse-to-fine manner. Li et al. \cite{li2022bridging,li2021deep} predict the trimap parallel to the alpha matte of the transition region and then fuse them to obtain the final alpha matte.

\section{Proposed Approach}

\begin{figure*}[t!]
\centerline{\includegraphics[width=0.95\textwidth]{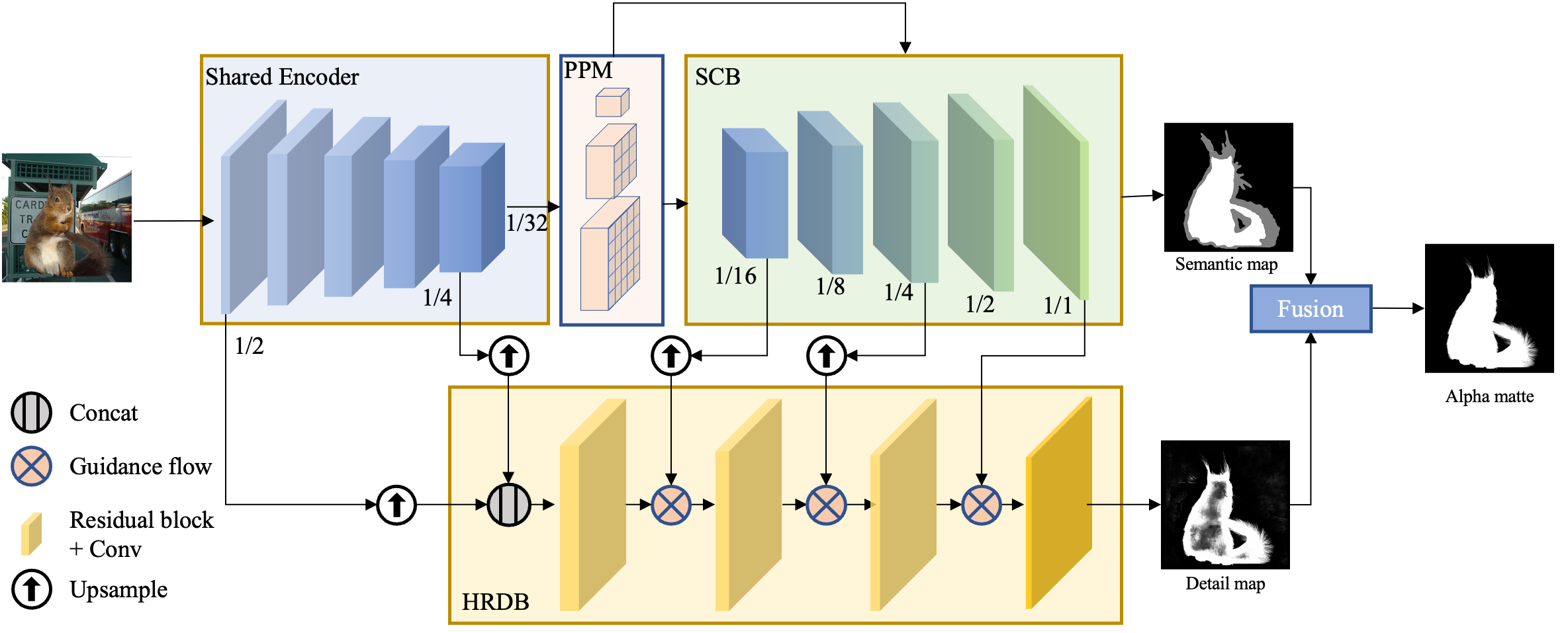}}
\caption{The overview of network architecture. HRNet is used as the encoder, the low-resolution output is used as the input of PPM, and the high-resolution output is used as the input of the high-resolution detail branch (HRDB). The semantic context branch (SCB) has five blocks, and the first, third, and fifth blocks guide the HRDB to learn semantic context. The HRDB maintains high-resolution inference to have a high-quality detail prediction.}
\label{fig:network}
\end{figure*}

Correct semantic context and clear details are two essential factors for high-accuracy image matting. In trimap-based approaches, the user-supplied trimap has already provided sufficient semantic context in advance, so that the network just focuses on the transition region. On the contrary, trimap-free approaches have no such prepared semantic context. Therefore, semantic context extraction is critical in a trimap-free method. As shown in Figure~\ref{fig:network}, we propose a semantic-aware network named PP-Matting, which consists of the semantic context branch (SCB) and the high-resolution detail branch (HRDB) to predict the semantic map and the detail map, respectively. To generate the final alpha matte, the semantic map and detail map are fused resulting in mutual enhancement. When a person looks at an image,  global context information is usually taken as meaningful guidance to help him dive into details. Inspired by the fact, we propose the design of guidance flow that achieves high-accuracy image matting with better extraction of object details.

\subsection{Network Architecture}
The proposed network consists of two branches, i.e. semantic context branch (SCB) and the high-resolution detail branch (HRDB), which share a common module as the encoder, i.e. shared encoder. The pyramid pooling module (PPM) has been proven to be effective in semantic segmentation, due to its powerful extraction of global context~\cite{zhao2017pyramid}. Thus, before SCB, we apply a PPM to strengthen semantic context. The guidance flow is used to connect HRDB and SCB, which is helpful for details prediction with correct semantic guidance. 

\textbf{Shared Encoder.}
Efficient feature extraction is a common part of any deep learning method. A matting task needs to combine high-level abstract semantic segmentation with high-resolution detail regression. Therefore, in addition to low-resolution feature representation, a high-resolution one is also necessary for image matting. In our work, we adopt HRNet48~\cite{wang2020deep} pre-trained on ImageNet as the shared encoder because of its powerful ability in high-resolution feature extraction. Rather than recovering high-resolution by upsampling, the shared encoder maintains high-resolution representation throughout the whole process and repeatedly fuses the multi-resolution representations to generate rich high-resolution features with strong position sensitivity. The encoder is shared between SCB and HRDB. 

\textbf{Semantic Context Branch.}
Without the trimap cues, previous matting approaches are hard to obtain correct semantic context and easily fall into local ambiguity. In this work, we design a separate semantic context branch (SCB) for global context extraction. The SCB consists of five blocks and the block consists of three ConvBNReLU (convolution, batch normalization (BN) and Rectified Linear Unit (ReLU)) and one bilinear upsample. Since high-level abstract features are important for a correct semantic map, we take 1/32 resolution outputs of the encoder as input of the SCB. Also, a PPM is used to strengthen semantic context further between the shared encoder and SCB. In addition, the output of SCB is supervised by a semantic segmentation task of three classes, i.e. foreground, background, and transition. As shown in Figure~\ref{fig:network}, the semantic map of SCB is very similar to a user-supplied trimap, which indicates the proposed model is capable of generating a reliable semantic-aware trimap automatically.

\textbf{High-Resolution Detail Branch.}
As shown in Figure~\ref{fig:seg_mat}, the degree of fine-grained details is one major difference between matting and segmentation. The details representation is usually kept in high-resolution feature maps. However, existing encoder-decoder segmentation networks with the downsampling-upsampling architecture do not maintain high-resolution representation efficiently. In this work, we propose the high-resolution detail branch (HRDB) to obtain fine-grained details in high resolution. Specifically, the HRDB consists of three residual blocks followed by a convolution layer. The initial input of HRDB is a combination of two intermediate features with corresponding upsampling scales, which contain low-level texture information and high-level abstract information, respectively. Between two blocks, guidance flow is used to obtain semantic context from SCB. The output of the HRDB is detail map that focuses on details representation in the transition region, and then it is fused with the semantic map in SCB to generate the final alpha matte. 

\begin{figure}[t]
\centering
    \includegraphics[width=0.3\textwidth]{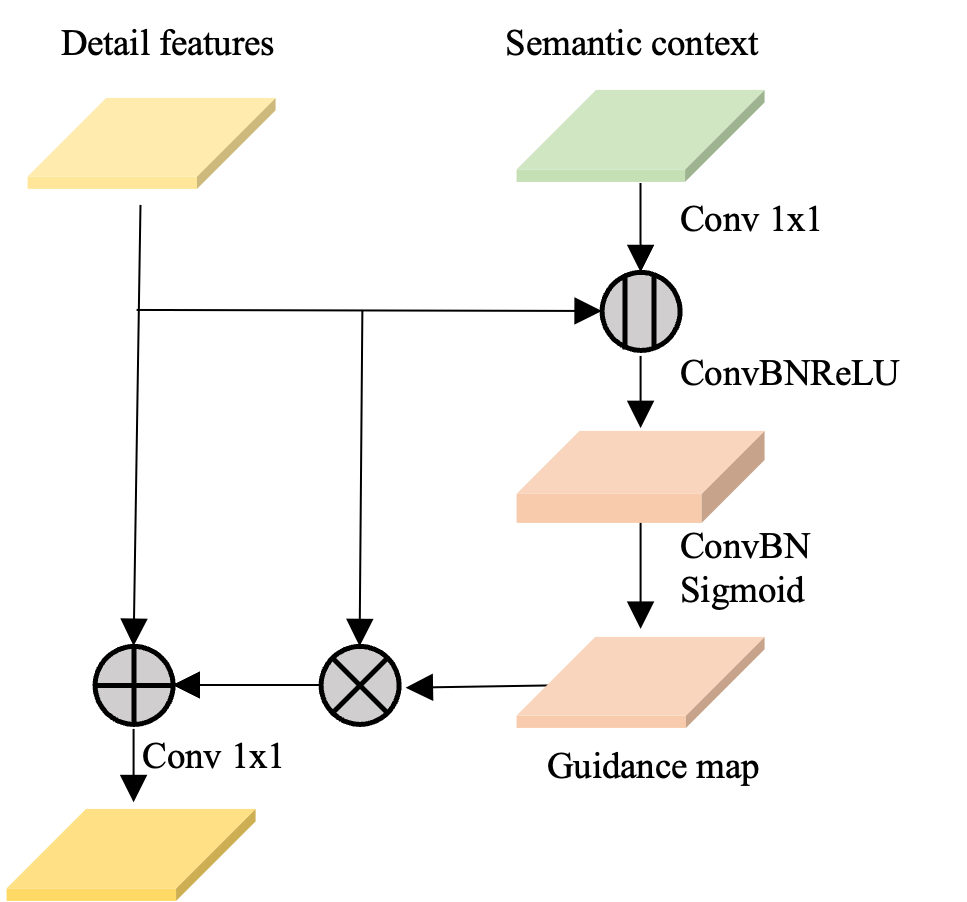}
    \caption{Guided flow architecture}
    \label{fig:gcl}
\end{figure}

\textbf{Guidance Flow.}
As we discussed above, the global semantic context provides meaningful guidance for details extraction. In this work, we apply the guidance flow strategy that guides semantic features of SCB to help details prediction in HRDB. Specifically, it is consisting of Gated Convolutional Layers (GCL)~\cite{takikawa2019gated}, which achieve the interaction between detail feature and semantic context as shown in Figure~\ref{fig:gcl}. In GCL, a guidance map $g \in R^{H \times W}$ is obtained as:
\begin{equation}
  g = \sigma(C_{1\times1}(s||d))
  \label{eq:attention}
\end{equation}
where $s$ and $d$ denote intermediate features of the SCB and HRDB, respectively. $||$ denotes a concatenation operation. $C_{1\times1}$ denotes the normalized $1\times1$ convolutional layer. $\sigma$ denotes the sigmoid function. The new detail feature map $\hat{d}$ is calculated as:
\begin{equation}
    \hat{d} = (d \odot g + d)^Tw
    \label{eq:gcl}
\end{equation}
where $\odot$ denotes an element-wise product, and $w$ denotes the channel-wise weighting kernel. In our method, we use three GCL modules with the first, third and last block of the SCB to guide the detail branch.

\subsection{Loss Function}

Our model leverages three losses. The first loss is a semantic loss in SCB denoted as $\mathcal{L}_{s}$, which is a cross-entropy loss of the 3-class segmentation task.
\begin{equation} \mathcal{L}_{s} = -\sum_{c=1}^{3}\sum_{i=1}^{\Omega}g_c^ilog(p_c^i) \label{eq:lt}
\end{equation}

where $c \in \{1,2,3\}$ denotes three classes in semantic map. $p_c^i \in [0,1]$ is the predicted probability for the c-th class at the i-th pixel, $g_c^i \in \{0,1\}$ is the corresponding ground truth. $\Omega$ denotes all pixels in the image.

\begin{equation} \Omega = \Omega^f \cup \Omega^b \cup \Omega^t\label{eq:set}
\end{equation}
where $\Omega^f$, $\Omega^b$, and $\Omega^t$ denote a set of foreground pixels, background pixels, and transition pixels, respectively.

The second loss is detail loss in HRDB denoted as $\mathcal{L}_{d}$. Following~\cite{xu2017deep, tang2019learning}, we sum alpha-prediction loss $\mathcal{L}_\alpha$ and the gradient loss $\mathcal{L}_{grad}$ as the detail loss. Since the HRDB is focusing on detail prediction, we only calculate it in the transition region.
\begin{equation} 
\mathcal{L}_{d} = \sum_{i=1}^{\Omega^t}(\mathcal{L}_\alpha^i(d) + \mathcal{L}_{grad}^i(d))
\label{eq:ld}
\end{equation}
The alpha-prediction loss is the absolute difference between the ground truth alpha values and the predicted alpha values at each pixel. For the differentiable property, it is formulated as follows.
\begin{equation} \mathcal{L}_\alpha^i(d) = \sqrt{(\alpha_d^i - \alpha_g^i)^2 + \epsilon^2} \label{eq:ldalpha}
\end{equation}
\begin{equation} \mathcal{L}_{grad}^i(d) = |\nabla\alpha_d^i - \nabla\alpha_g^i| \label{eq:ldgrad}
\end{equation}
where $\alpha_{d}^i$ is the detail map value at i-th pixel, and $\alpha_g^i$ is the ground-truth alpha value. $\epsilon$ is a small value which is set to $10^{-6}$ in our experiment. $\nabla$ denotes the calculation of the gradient magnitude.

The third loss is fusion loss in final alpha matte denoted as $\mathcal{L}_f$ which consists of alpha-prediction loss, gradient loss and composition loss as follow.
\begin{equation}
    \mathcal{L}_{f} = \sum_i^\Omega(\mathcal{L}_{\alpha}^i(p) + \mathcal{L}_{grad}^i(p) + \mathcal{L}_{comp}^i)
    \label{eq:lcm}
\end{equation}
where $\mathcal{L}_{comp}^i$ is the absolute difference between the ground truth RGB colors and the predicted RGB colors composited by the ground truth foreground, the ground truth background and the predicted alpha matte.
\begin{equation}
    \mathcal{L}_{comp}^i = \sqrt{(I_p^i - I_g^i)^2 + \epsilon^2}
    \label{eq:lcmcomp}
\end{equation}
\begin{equation}
  I_p^i = \alpha_p^i F_g^i + (1 - \alpha_p^i)B_g^i
  \label{eq:important}
\end{equation}

where $I_p$ denotes the image composited by the predicted alpha, and $I_g$ denotes the ground truth image. $\epsilon$ is a small value which is set to $10^{-6}$ in our experiment. $\mathcal{L}_{\alpha}^i(p)$ and $\mathcal{L}_{grad}^i(p)$ are the same to Eq.\ref{eq:ldalpha}, \ref{eq:ldgrad}, but the predicted alpha values are final alpha matte.

The final weighted loss is calculated as follows.
\begin{equation} \mathcal{L} = \lambda_1\mathcal{L}_{s} + \lambda_2\mathcal{L}_{d} + \lambda_3\mathcal{L}_{f} \label{eq:loss}
\end{equation}

\section{Experiments}

\subsection{Datasets}
We conduct the experiments on two public datasets: Distinctions-646~\cite{qiao2020attention} and Adobe Composition-1k~\cite{xu2017deep}. In Distinctions-646, the training set contains 596 foreground objects with their corresponding ground truth alpha mattes. The test set contains 50 foreground images and their corresponding alpha mattes. In Adobe Composition-1k, there are 431 foreground images in the training set and 50 in the test set. For the two datasets, each foreground image is combined with 100 background images for training and 20 background images for testing, which is the same as~\cite{xu2017deep}.

\subsection{Implementation Details}
\label{sec:imp}

\textbf{Training settings.} In the training phase, input images are randomly cropped to $512\times512$, $640\times640$ and $800\times800$. Note that if the size of an image is smaller than 512, it is padded to $512\times512$ without cropping. Then they are resized to a resolution of $512\times512$ and augmented by random distortion, random blurring, and random horizontal flipping. We apply stochastic gradient descent (SGD) optimizer with the momentum of 0.9 and weight decay of $4e^{-5}$. The learning rate is initialized to 0.01, and adjusted by the poly policy with the power of 0.9 and 300k iterations. The coefficients in eq.\ref{eq:loss} are set to $\lambda_1 = \lambda_2 = \lambda_3 = 1.0$. All of our experiments are conducted on a single Tesla V100 GPU with a batch size of 4 using PaddlePaddle\footnote{https://github.com/PaddlePaddle/Paddle}~\cite{ma2019paddlepaddle}. Code and pretrained models will be available at PaddleSeg\footnote{https://github.com/PaddlePaddle/PaddleSeg}~\cite{liu2021paddleseg}.

\textbf{Evaluation metrics.} The alpha mattes are evaluated by four common quantitative metrics: the sum of absolute differences (SAD), mean squared error (MSE), gradient (Grad), and connectivity (Conn) proposed by~\cite{rhemann2009perceptually}. The lower value of the metric, The better the prediction quality.

\subsection{Evaluation Results}

\textbf{Distinctions-646 testing dataset.} We evaluate the proposed PP-Matting with 8 previous matting methods on Disctinctions-646 testing dataset. The quantitative comparison results are given in Table~\ref{tab:distinctions}. Our method has much better results in all four metrics than the top-performing traditional methods, e.g., ClosedForm~\cite{levin2007closed}, KNN Matting~\cite{chen2013knn}. Compared with DIM, a trimap-based deep learning method, our method also shows comparable performance. Among the 4 metrics, SAD and Conn are better than DIM, and only the Grad metric is slightly worse. All methods at the upper part of the table require an RGB image with a user-supplied trimap, while our method only takes a single image to generate the alpha matte. In addition, we also compare PP-Matting with a recent state-of-the-art method, HAttMatting~\cite{qiao2020attention}. Note that HAttMatting is not open-sourced, so we refer to the results in their paper directly. PP-Matting is better than HAttMatting on SAD and Conn metrics, while slightly worse than it on Grad.

\begin{table}
  \centering
  \begin{tabular}{@{}lcccc@{}}
    \toprule
    Methods & SAD$\downarrow$ & MSE$\downarrow$ & Grad$\downarrow$ & Conn$\downarrow$ \\
    \midrule
    Share Matting \cite{gastal2010shared} & 119.56 & 0.026 & 129.61 & 114.37 \\
    Global Matting\cite{he2011global} & 135.56 & 0.039 & 119.53 & 136.44 \\
    ClosedForm \cite{levin2007closed} & 105.73 & 0.023 & 91.76 & 114.55 \\
    KNN Matting \cite{chen2013knn} & 116.68 & 0.025 & 103.15 & 121.45 \\
    DCNN \cite{shen2016deep} & 103.81 & 0.020 & 82.45 & 99.96 \\
    Info-Flow \cite{aksoy2017designing} & 78.89 & 0.016 & 58.72 & 80.47 \\
    DIM \cite{xu2017deep} & \bf{47.56} & \bf{0.009} & \bf{43.29} & \bf{55.90} \\
    \midrule
    HAttMatting \cite{qiao2020attention} & 48.98 & 0.009 & \bf{41.57} & 49.93 \\
    PP-Matting & \bf{40.69} & \bf{0.009} & 43.91 & \bf{40.56} \\
    \bottomrule
  \end{tabular}
  \caption{The quantitative comparison on Distinctions-646 testing set. Upper part: trimap-based approaches. Lower part: trimap-free approaches.}
  \label{tab:distinctions}
\end{table}

\textbf{Composition-1k testing dataset.} On the Composition-1k testing dataset, we evaluate PP-Matting with 13 previous methods including more deep-learning methods. The quantitative comparison results are given in Table~\ref{tab:composition}. Our method has much better results in all four metrics than the traditional methods and trimap-based deep learning methods, i.e., DIM~\cite{xu2017deep}, AlphaGAN~\cite{lutz2018alphagan}, SampleNet~\cite{tang2019learning}. Trimap-based methods can explicitly obtain stronger semantic context with the trimap input, while our method also achieves better results through using the semantic context branch. For trimap-free methods, PP-Matting is better than Late Fusion~\cite{zhang2019late} and HAttMatting on all metrics except SAD. The results demonstrate the superiority of PP-Matting over previous methods.

\begin{table}
  \centering
  \begin{tabular}{@{}lcccc@{}}
    \toprule
    Methods & SAD$\downarrow$ & MSE$\downarrow$ & Grad$\downarrow$ & Conn$\downarrow$ \\
    \midrule
    Share Matting \cite{gastal2010shared} & 125.37 & 0.029 & 144.28 & 123.53 \\
    Global Matting\cite{he2011global} & 156.88 & 0.042 & 112.28 & 155.08 \\
    ClosedForm \cite{levin2007closed} & 124.68 & 0.025 & 115.31 & 106.06 \\
    KNN Matting \cite{chen2013knn} & 126.24 & 0.025 & 117.17 & 131.05 \\
    DCNN \cite{shen2016deep} & 115.82 & 0.023 & 107.36 & 111.23 \\
    Info-Flow \cite{aksoy2017designing} & 70.36 & 0.013 & 42.79 & 70.66 \\
    DIM \cite{xu2017deep} & 48.87 & 0.008 & 31.04 & 50.36 \\
    AlphaGAN \cite{lutz2018alphagan} & 90.94 & 0.018 & 93.92& 95.29 \\
    SampleNet \cite{tang2019learning} & 48.03 & 0.008 & 35.19 & 56.55 \\
    IndexNet \cite{lu2019indices} & \bf{44.52} & \bf{0.005} & \bf{29.88} & \bf{42.37} \\
    \midrule
    Late Fusion \cite{zhang2019late} & 58.34 & 0.011 & 41.63 & 59.74 \\
    HAttMatting \cite{qiao2020attention} & \bf{44.01} & 0.007 & 29.26 & 46.41 \\
    PP-Matting & 46.22 & \bf{0.005} & \bf{22.69} & \bf{45.40} \\
    \bottomrule
  \end{tabular}
  \caption{The quantitative comparison on Composition-1k testing set. Upper part: trimap-based approaches. Lower part: trimap-free approaches.}
  \label{tab:composition}
\end{table}

\subsection{Ablation Study}
In PP-Matting, the guidance flow of semantic context is a core design. To illustrate its effectiveness, we conduct ablation experiments on Distinctions-646. The experiment settings are the same as section~\ref{sec:imp} , but we reduce the number of training iters to 100k. As the results are shown in Table~\ref{tab:flow ablation}, the design of guidance flow can improve the performance significantly. We can find that even using only one block in SCB benefits the SAD, Grad, and Conn metrics. In our experiment, we take the first, the third, and the fifth block in SCB as guidance flow, because there is no obvious benefit from continuing to increase the number of blocks.

When we get the semantic map and detail map, we need to fuse them to the final alpha matte. In our experiments, we replace the transition region in the semantic map with the results of the detail map, which is denoted as Rep FM. There, we explore the other two methods. The first one is that we only use the detail map as the final alpha matte directly, denoted as w/o FM. The second one is that we concatenate the detail map and semantic map, and then use a 1x1 convolution layer to fuse it, denoted as Conv FM. The results in Table~\ref{tab:fusion-ablation} show that the replacement method is more effective without extra computation overhead.

\begin{table}
  \centering
  \begin{tabular}{@{}lcccc@{}}
    \toprule
    Methods & SAD$\downarrow$ & MSE$\downarrow$ & Grad$\downarrow$ & Conn$\downarrow$ \\
    \midrule
    w/o GF & 52.52 & 0.0118 & 54.22 & 53.42 \\
    5 & 51.07 & 0.0119 & 53.12 & 51.88  \\
    1, 3, 5 &50.79 & \bf{0.0113} & \bf{52.99} & 51.40 \\
    1, 2, 3, 4, 5 & \bf{49.55} & 0.0116 & 53.20 & \bf{50.67} \\
    \bottomrule
  \end{tabular}
  \caption{Ablation study on guidance flow. w/o GF: without guidance flow between SCB and HRDB. The number 5 means that using the first block in the SCB as guidance flow. The number 1, 3, 5 means using the first, the third, and the fifth block in SCB as guidance flow.}
  \label{tab:flow ablation}
\end{table}

\begin{table}
  \centering
  \begin{tabular}{@{}lcccc@{}}
    \toprule
    Methods & SAD$\downarrow$ & MSE$\downarrow$ & Grad$\downarrow$ & Conn$\downarrow$ \\
    \midrule
    w/o FM & 58.71	& 0.0156 & 67.81 & 60.58 \\
    Conv FM & 51.82 & 0.0124 & 53.39 & 53.16 \\
    Rep FM &\bf{50.79} & \bf{0.0113} & \bf{52.99} & \bf{51.40} \\
    \bottomrule
  \end{tabular}
  \caption{Ablation study on fusion module. w/o FM: use the HRDB to predict the final alpha matte. Conv FM: use the convolution module to fuse the sematic map and the detail map. Rep FM: replace the transition region in semantic map with results of detail map.}
  \label{tab:fusion-ablation}
\end{table}

\begin{figure*}
    \centering
    \begin{subfigure}{0.19\linewidth}
        \centerline{\includegraphics[width=0.95\textwidth]{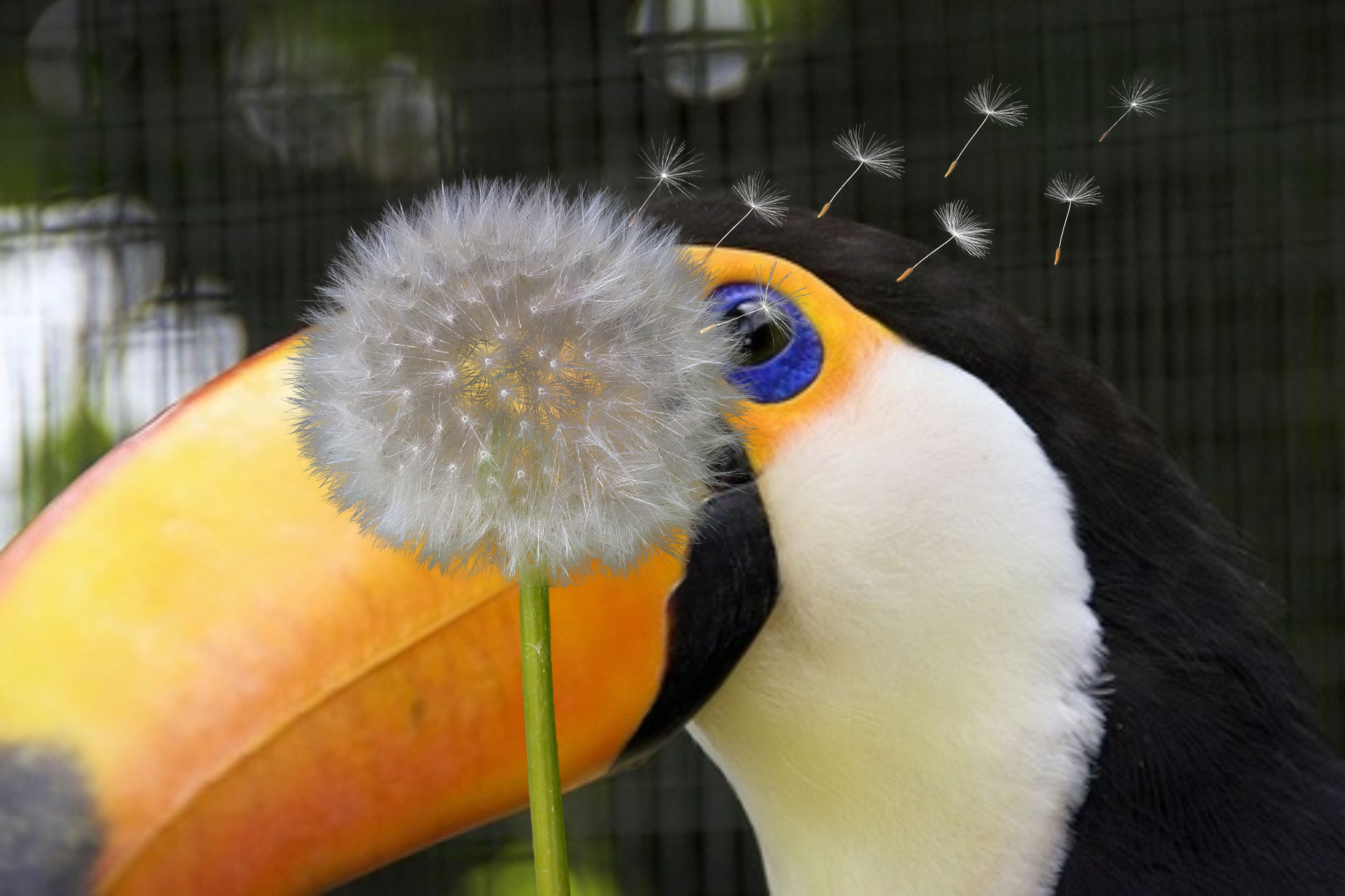}}
        \caption{Image}
        \label{fig:test_11_image}
    \end{subfigure}
    \centering
    \begin{subfigure}{0.19\linewidth}
        \centerline{\includegraphics[width=0.95\textwidth]{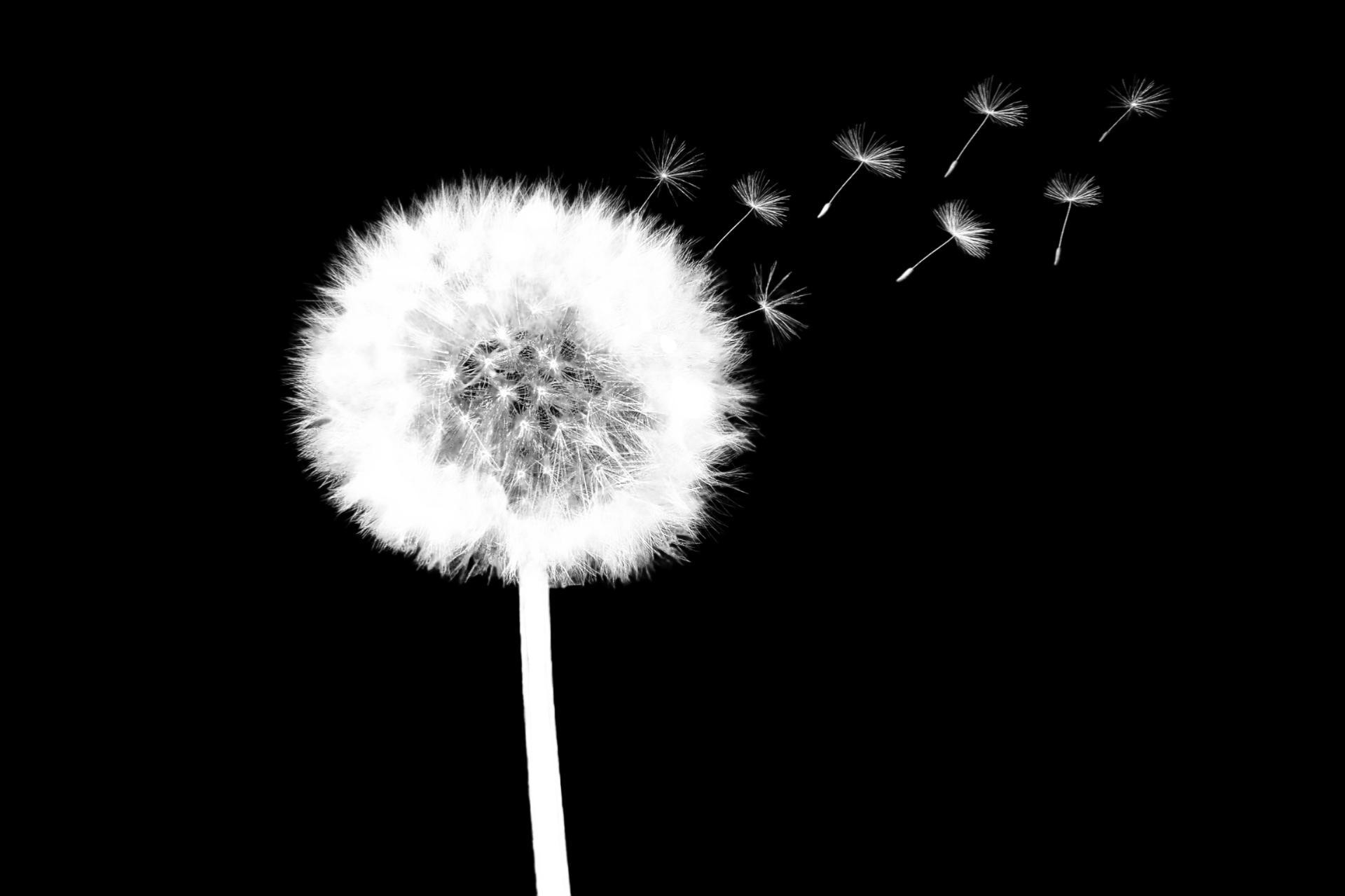}}
        \caption{Alpha matte}
        \label{fig:test_11_image}
    \end{subfigure}
    \centering
    \begin{subfigure}{0.19\linewidth}
        \centerline{\includegraphics[width=0.95\textwidth]{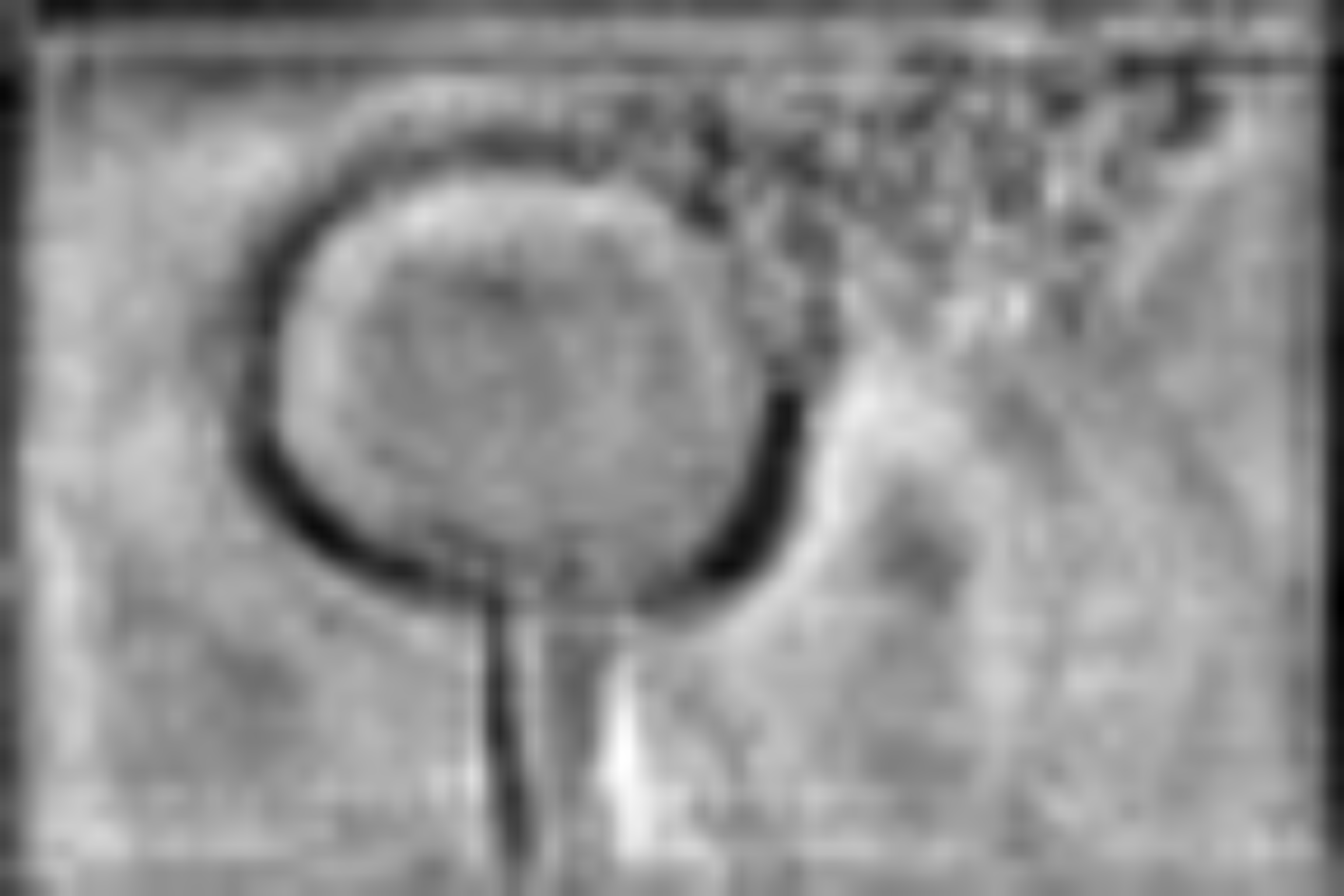}}
        \caption{s1}
        \label{fig:test_11_s1}
    \end{subfigure}
    \centering
    \begin{subfigure}{0.19\linewidth}
        \centerline{\includegraphics[width=0.95\textwidth]{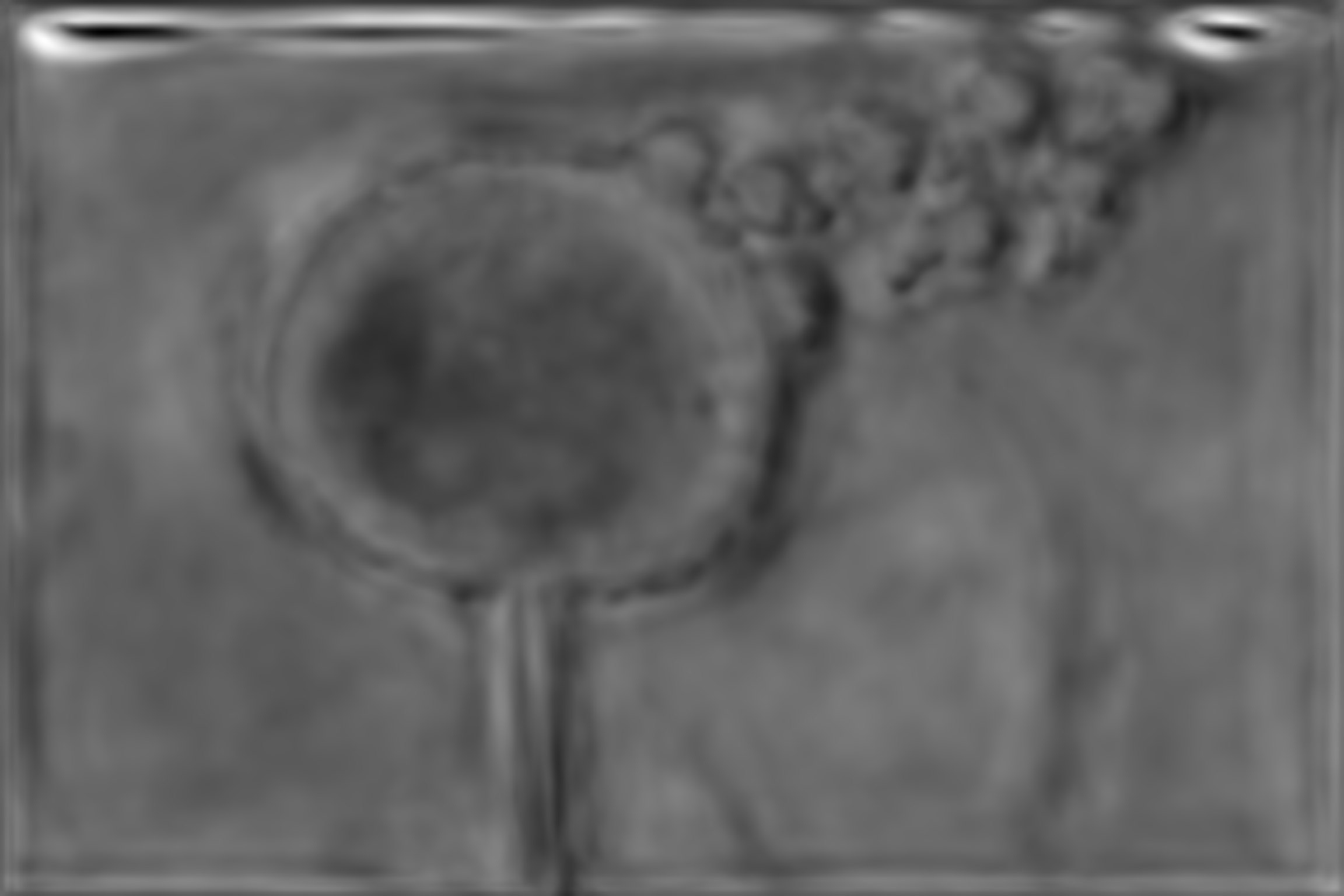}}
        \caption{s2}
        \label{fig:test_11_s2}
    \end{subfigure}
    \centering
    \begin{subfigure}{0.19\linewidth}
        \centerline{\includegraphics[width=0.95\textwidth]{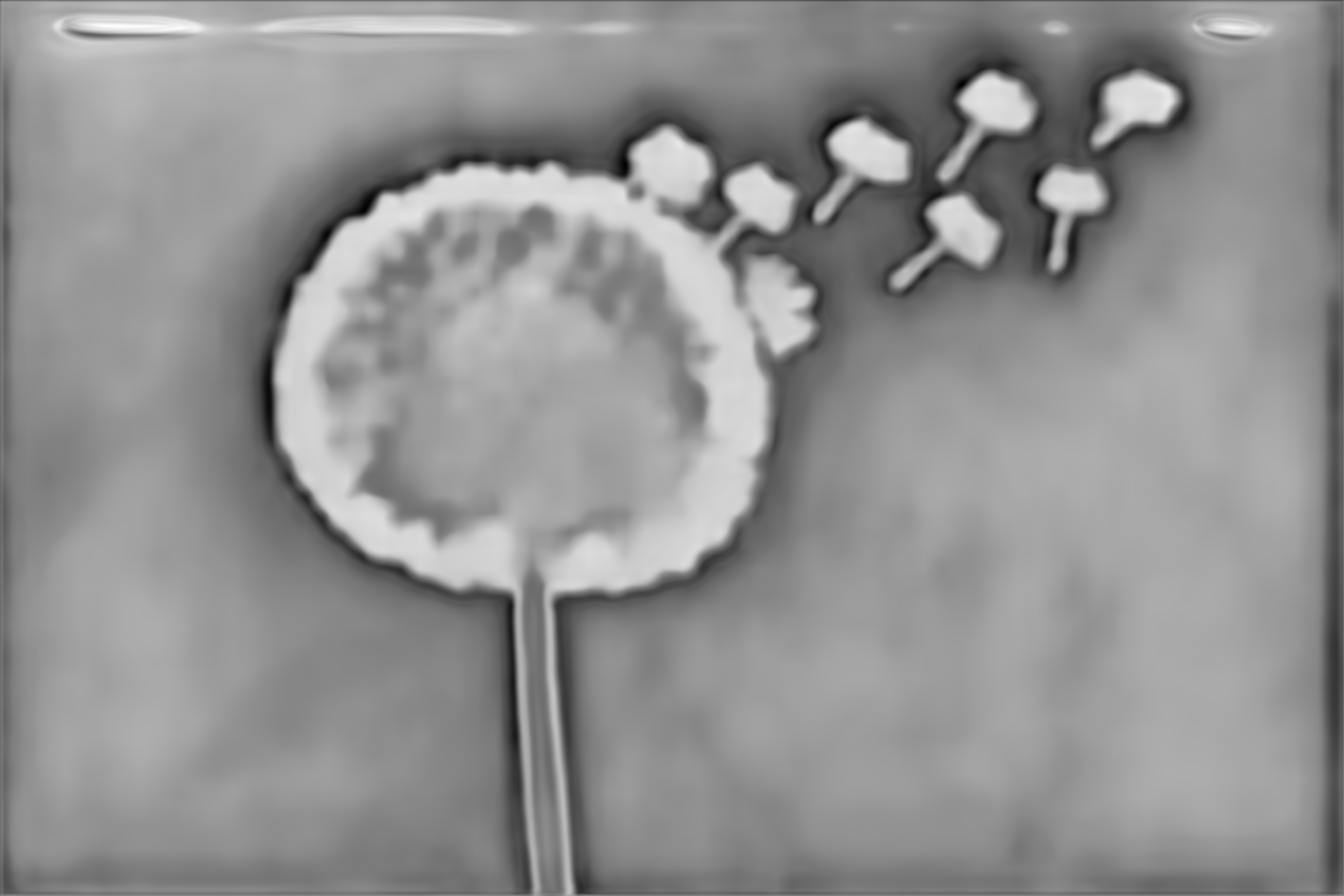}}
        \caption{s3}
        \label{fig:test_11_s3}
    \end{subfigure}
    \centering
    \begin{subfigure}{0.19\linewidth}
        \centerline{\includegraphics[width=0.95\textwidth]{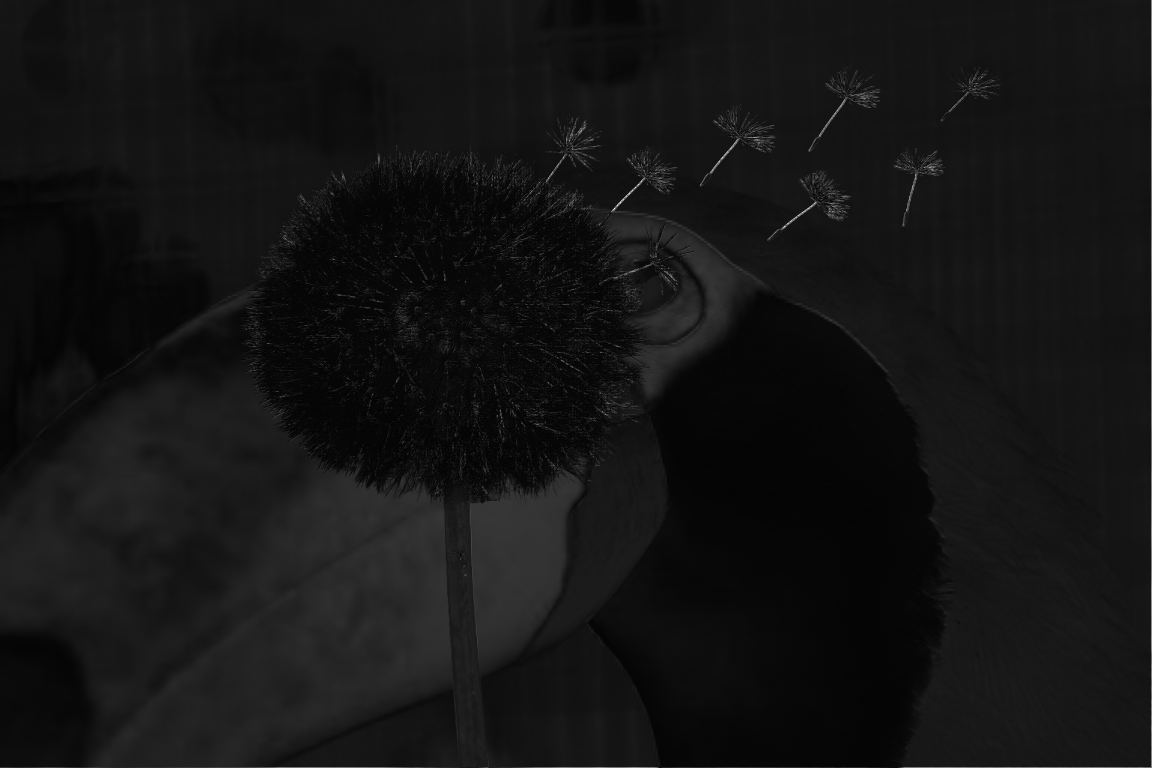}}
        \caption{Low-level feature}
        \label{fig:test_11_low_level_fea}
    \end{subfigure}
    \centering
    \begin{subfigure}{0.19\linewidth}
        \centerline{\includegraphics[width=0.95\textwidth]{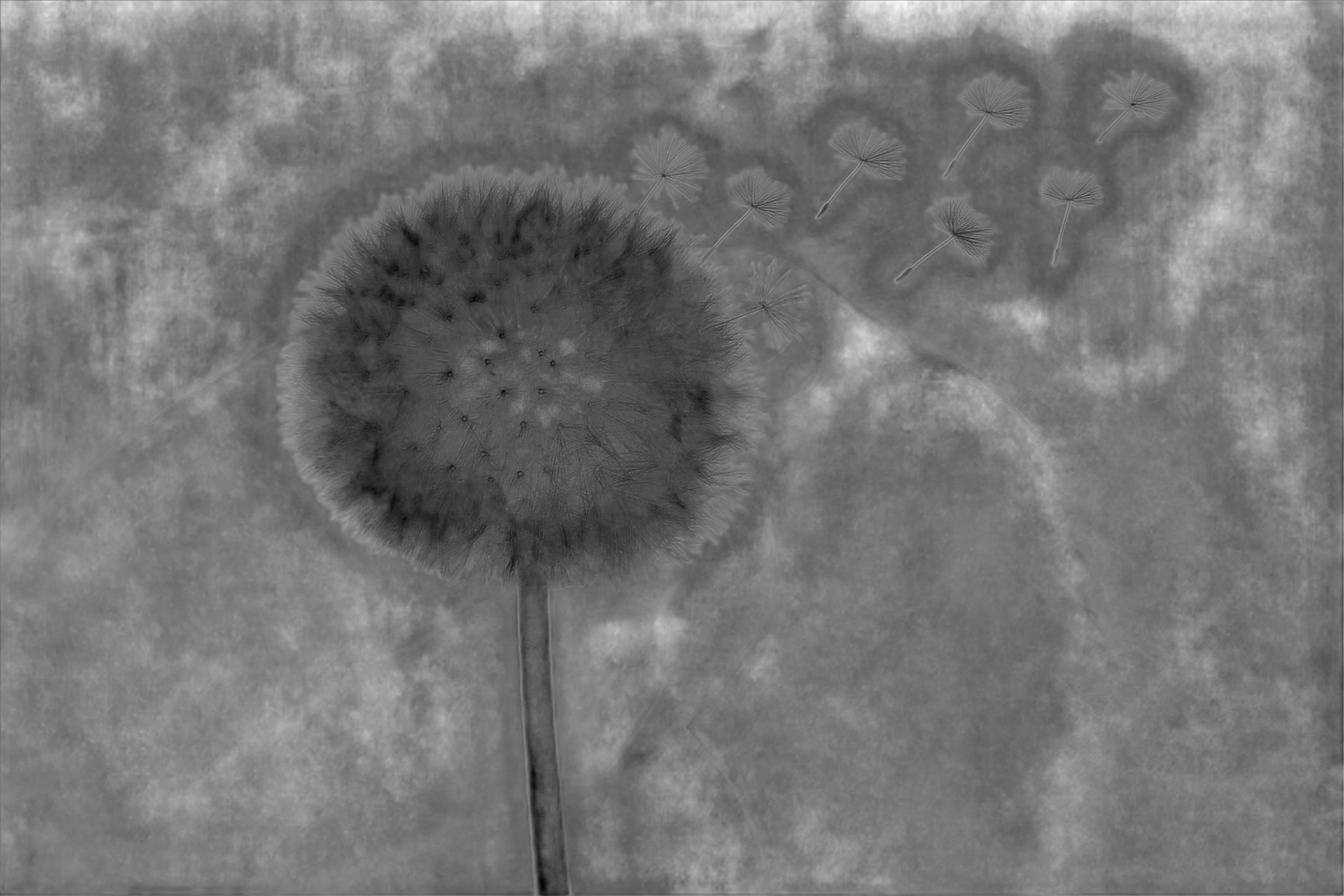}}
        \caption{g1}
        \label{fig:test_11_g1}
    \end{subfigure}
    \centering
    \begin{subfigure}{0.19\linewidth}
        \centerline{\includegraphics[width=0.95\textwidth]{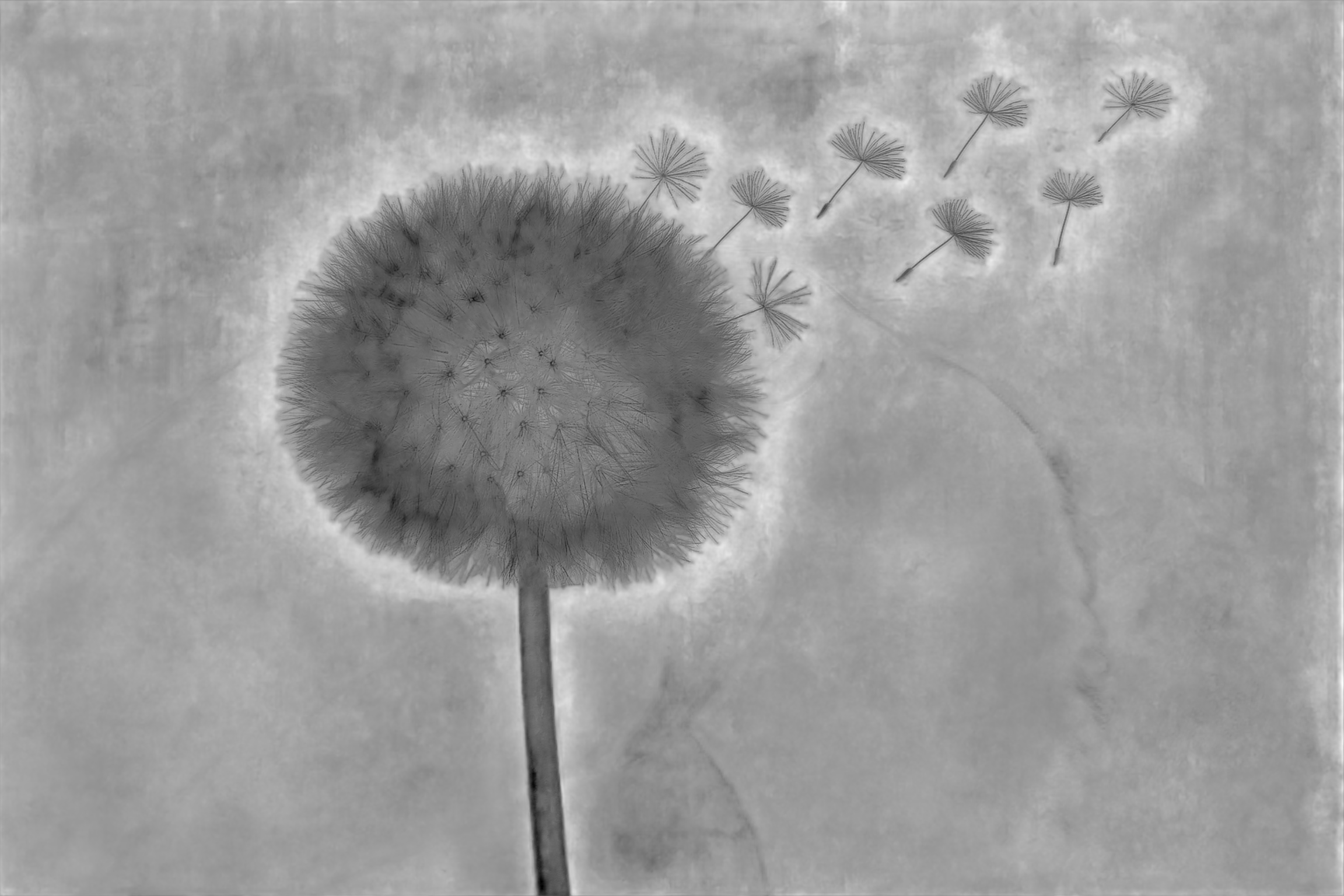}}
        \caption{g2}
        \label{fig:test_11_g2}
    \end{subfigure}
    \centering
    \begin{subfigure}{0.19\linewidth}
        \centerline{\includegraphics[width=0.95\textwidth]{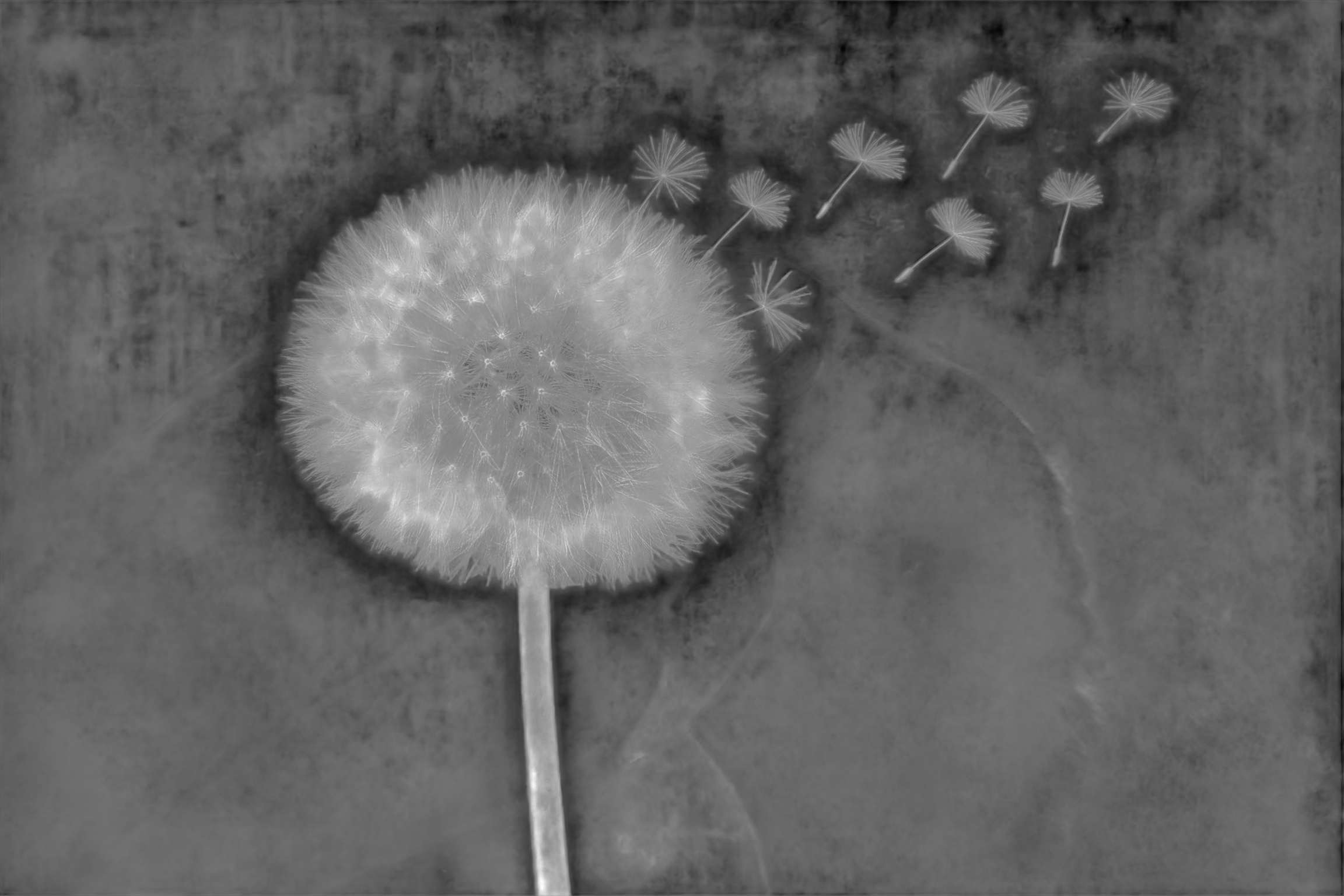}}
        \caption{g3}
        \label{fig:test_11_g3}
    \end{subfigure}
    \centering
    \begin{subfigure}{0.19\linewidth}
        \centerline{\includegraphics[width=0.95\textwidth]{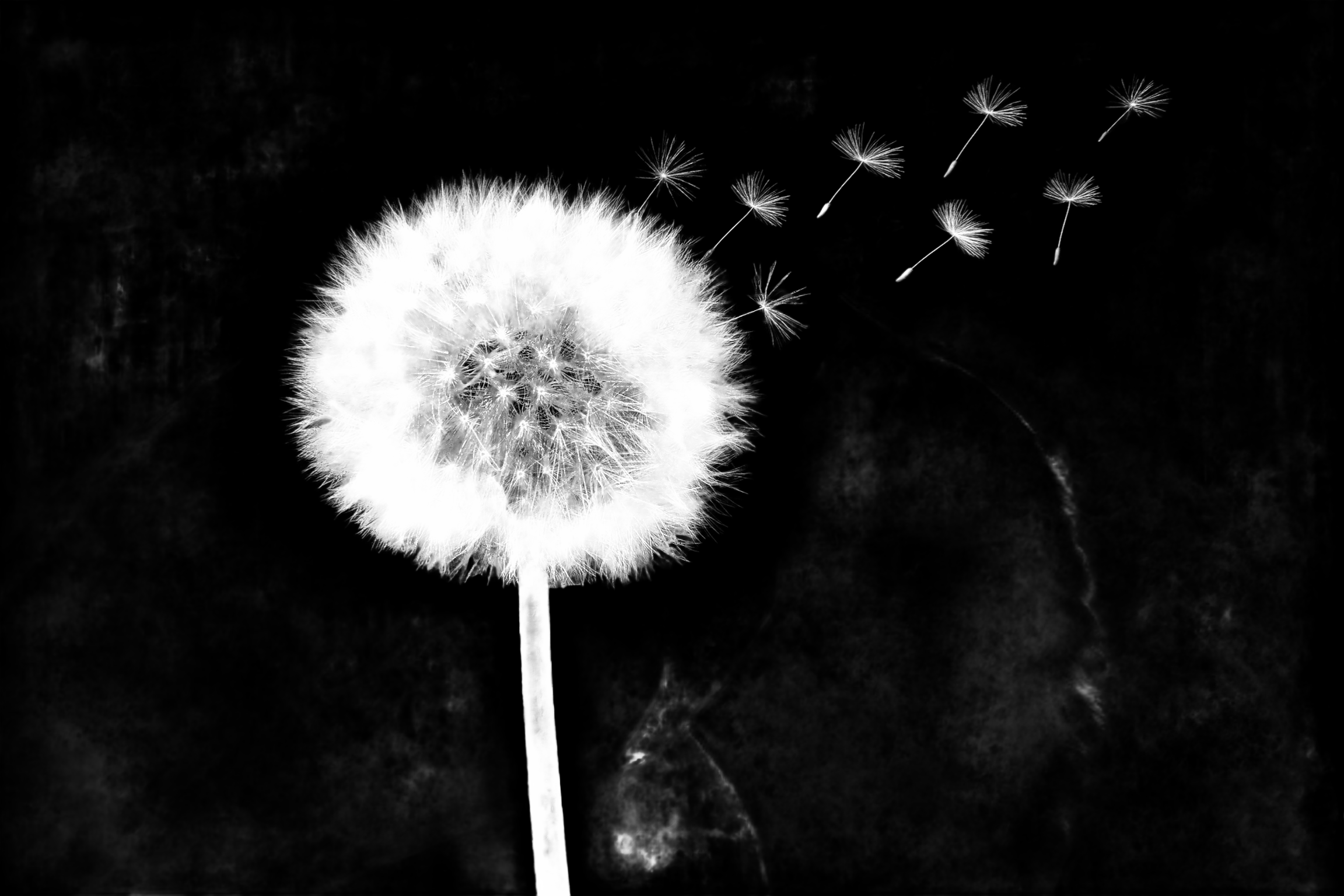}}
        \caption{Detail map}
        \label{fig:test_11_detail}
    \end{subfigure}

\caption{Visualization of the features and results. Alpha matte is the final prediction result. s1, s2, and s3 are the feature maps from the SCB. Low-level feature is from the first block of the encoder. g1, g2, and g3 are the output of GCLs. Detail map is the output of HRDB.}
\label{fig:visual_fea}
\end{figure*}

To further analyze the role of the guidance flow, we visualize the intermediate features of the network. As shown in figure \ref{fig:visual_fea}, the s1, s2 and s3 are features from the first, third, and fifth block in the SCB. The g1, g2, and g3 are the first channel feature of the GCLs output in the HRDB. From s1 to s3, we can see that the semantic map specifies the transition region clearly, which guides the HRDB to focus on the detail and texture in the region and ignore the background and foreground as shown in g1, g2, and g3. Finally, the HRDB gets the high-accuracy detail map in the transition region.

\begin{figure*}
    \centering
    \begin{subfigure}{0.19\linewidth}
        \centerline{\includegraphics[width=0.95\textwidth]{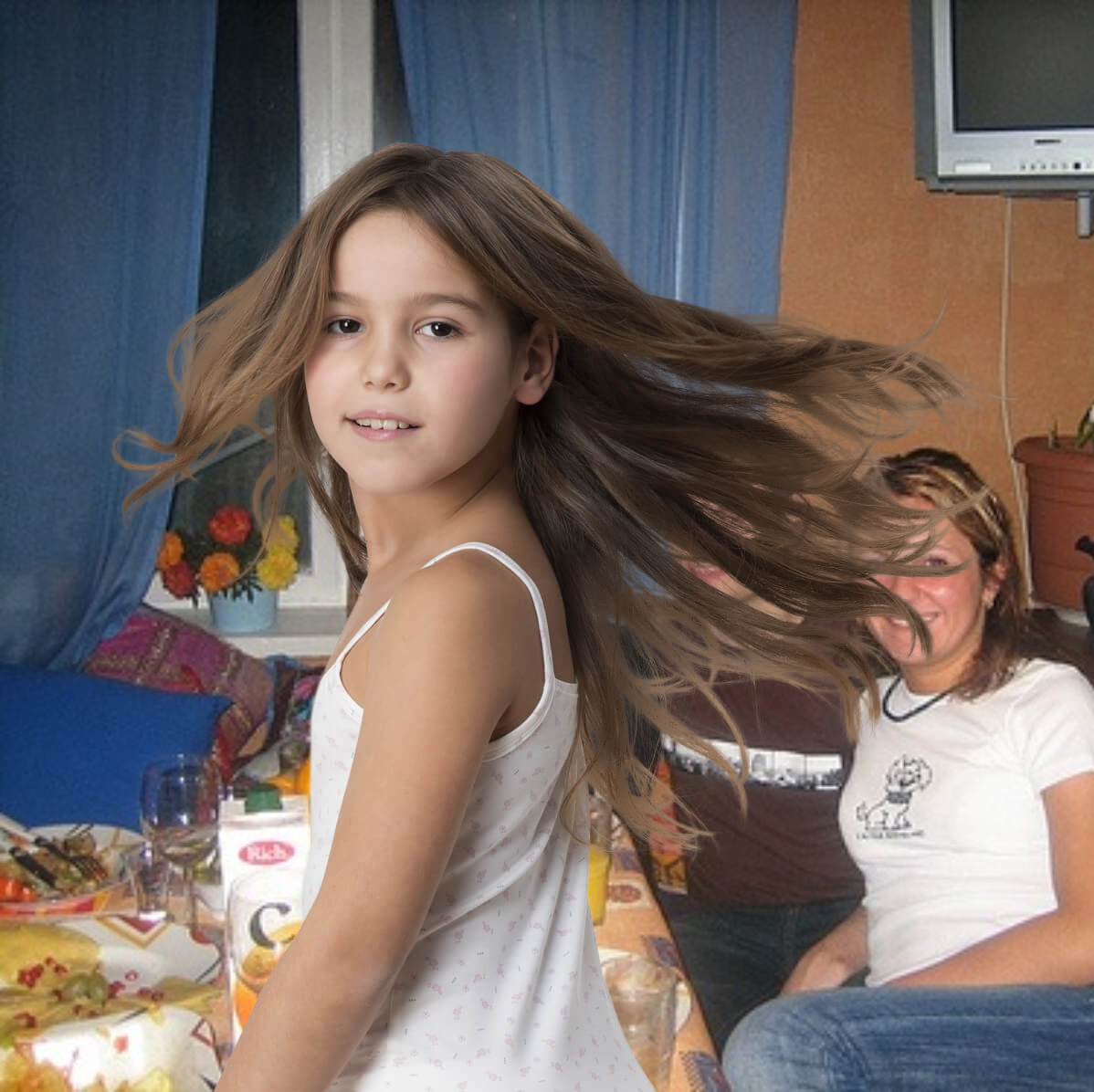}}
        \caption{Image}
        \label{fig:test_4_image}
    \end{subfigure}
    \centering
    \begin{subfigure}{0.19\linewidth}
        \centerline{\includegraphics[width=0.95\textwidth]{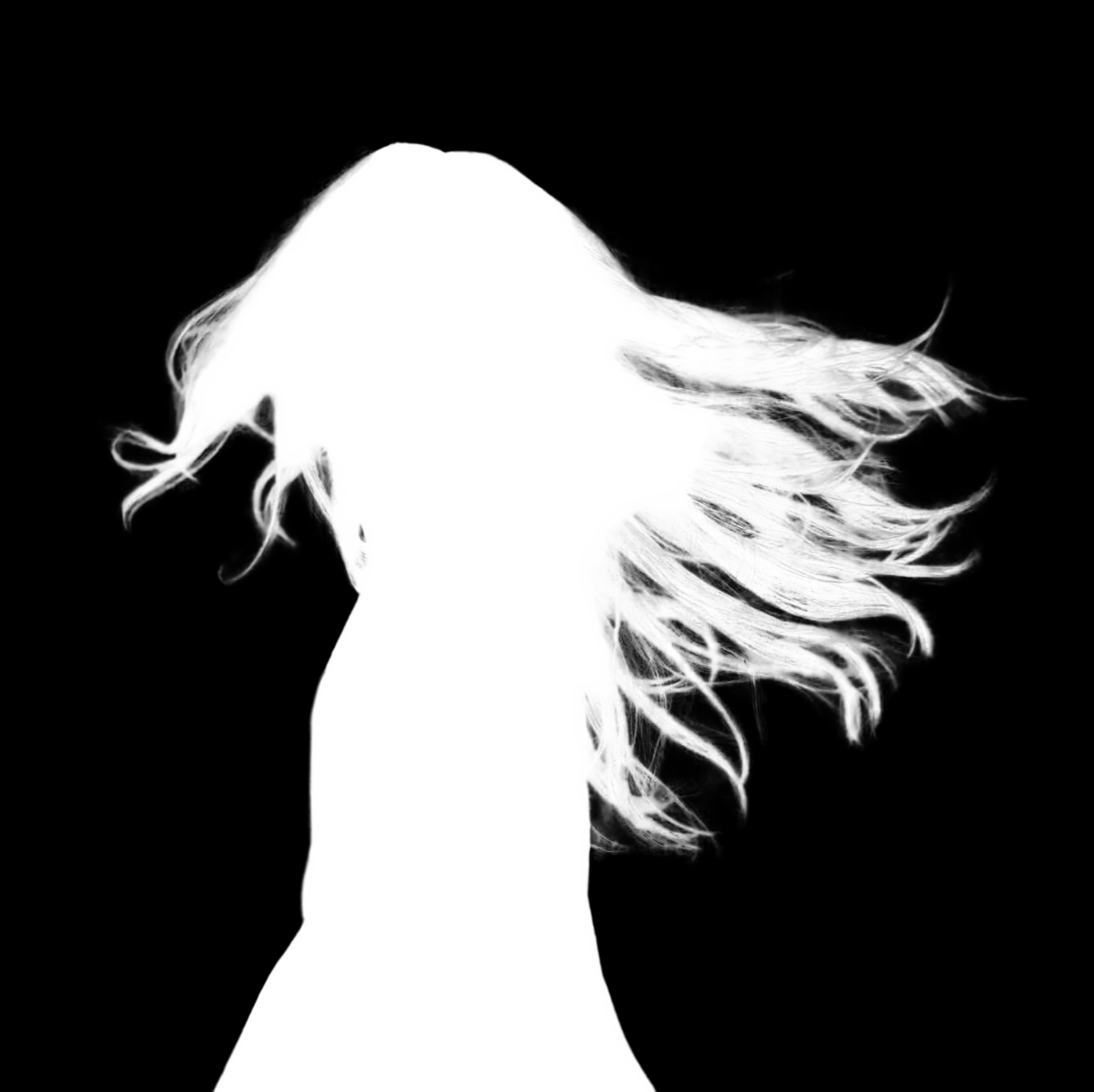}}
        \caption{Alpha matte}
        \label{fig:test_4_image}
    \end{subfigure}
    \centering
    \begin{subfigure}{0.19\linewidth}
        \centerline{\includegraphics[width=0.95\textwidth]{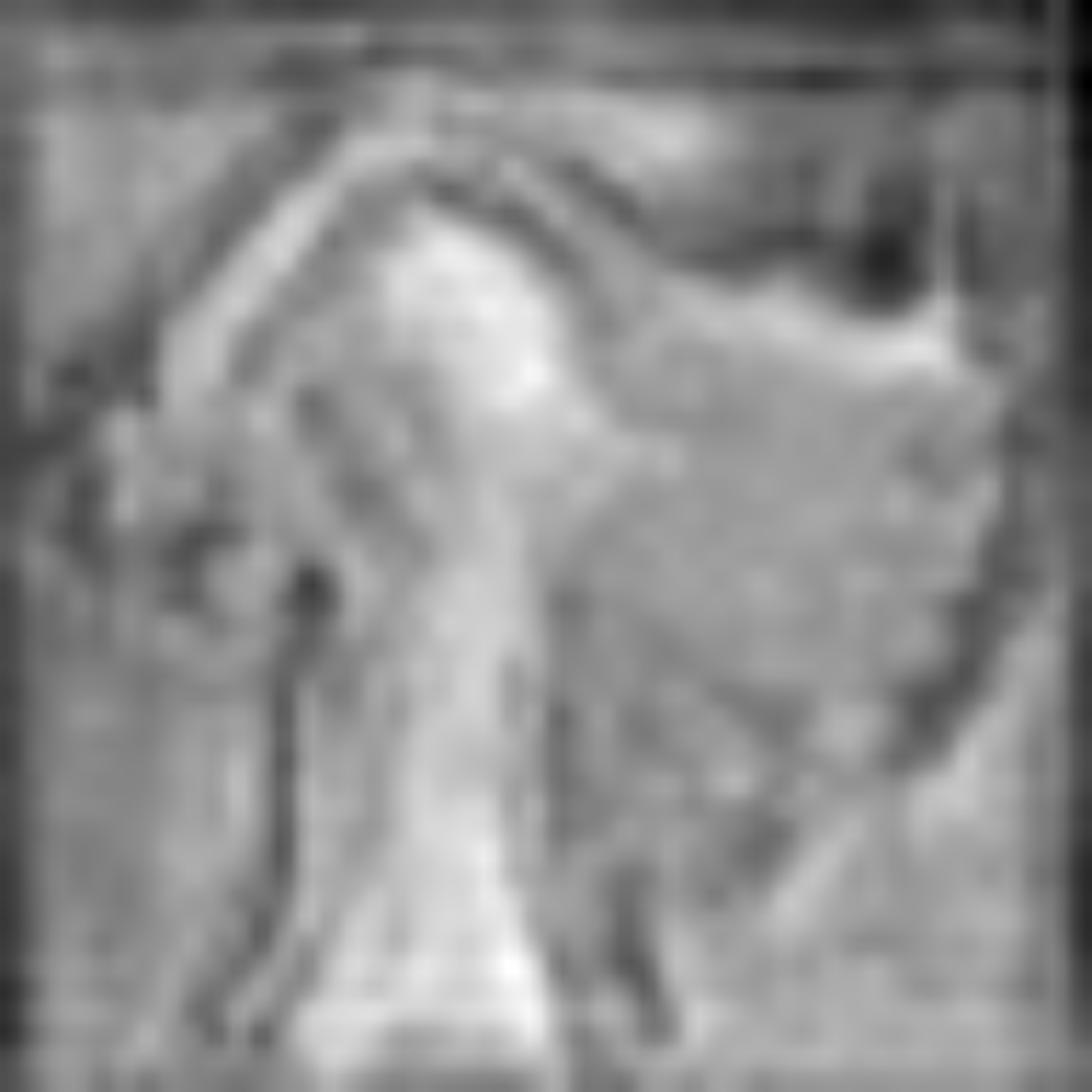}}
        \caption{s1}
        \label{fig:test_4_s1}
    \end{subfigure}
    \centering
    \begin{subfigure}{0.19\linewidth}
        \centerline{\includegraphics[width=0.95\textwidth]{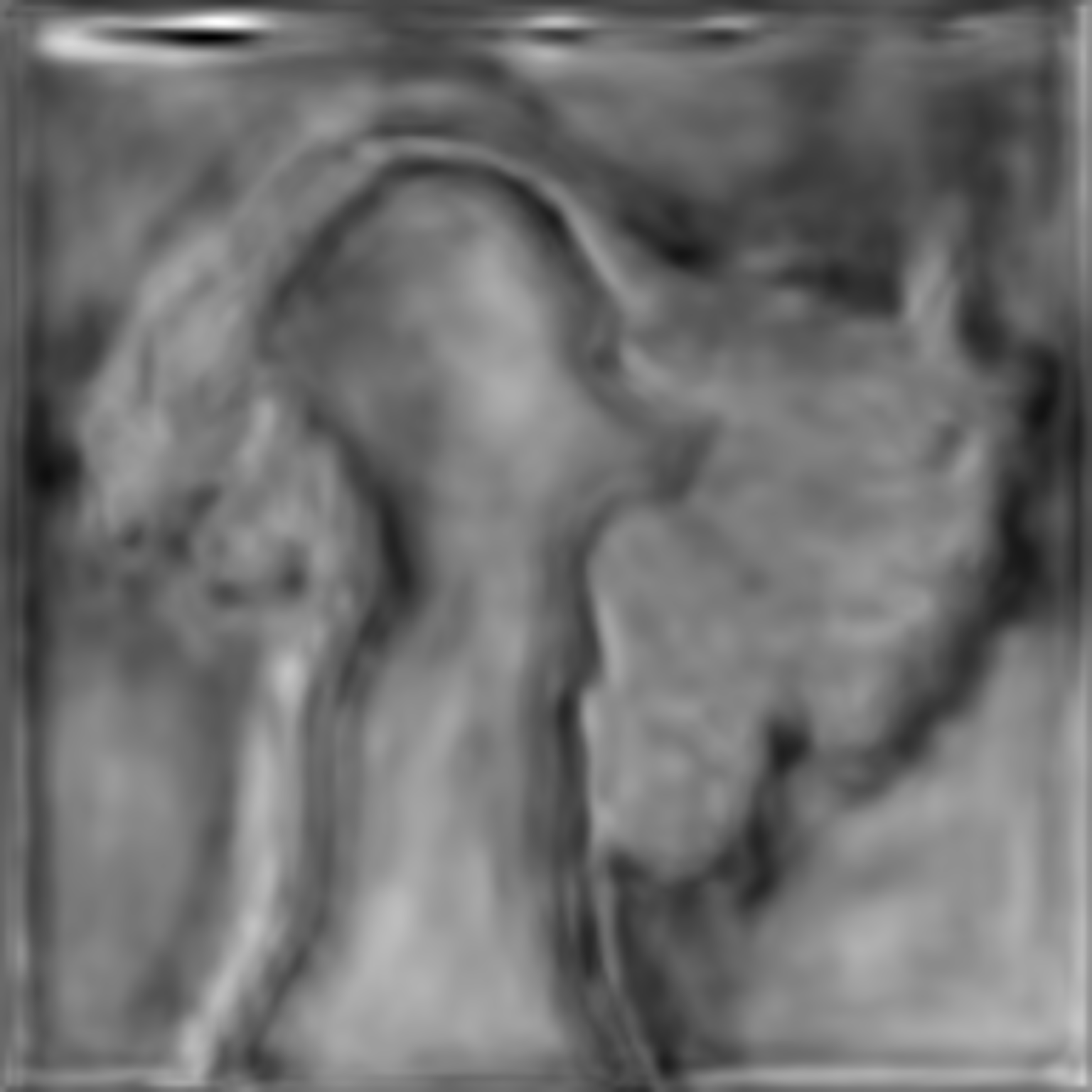}}
        \caption{s2}
        \label{fig:test_4_s2}
    \end{subfigure}
    \centering
    \begin{subfigure}{0.19\linewidth}
        \centerline{\includegraphics[width=0.95\textwidth]{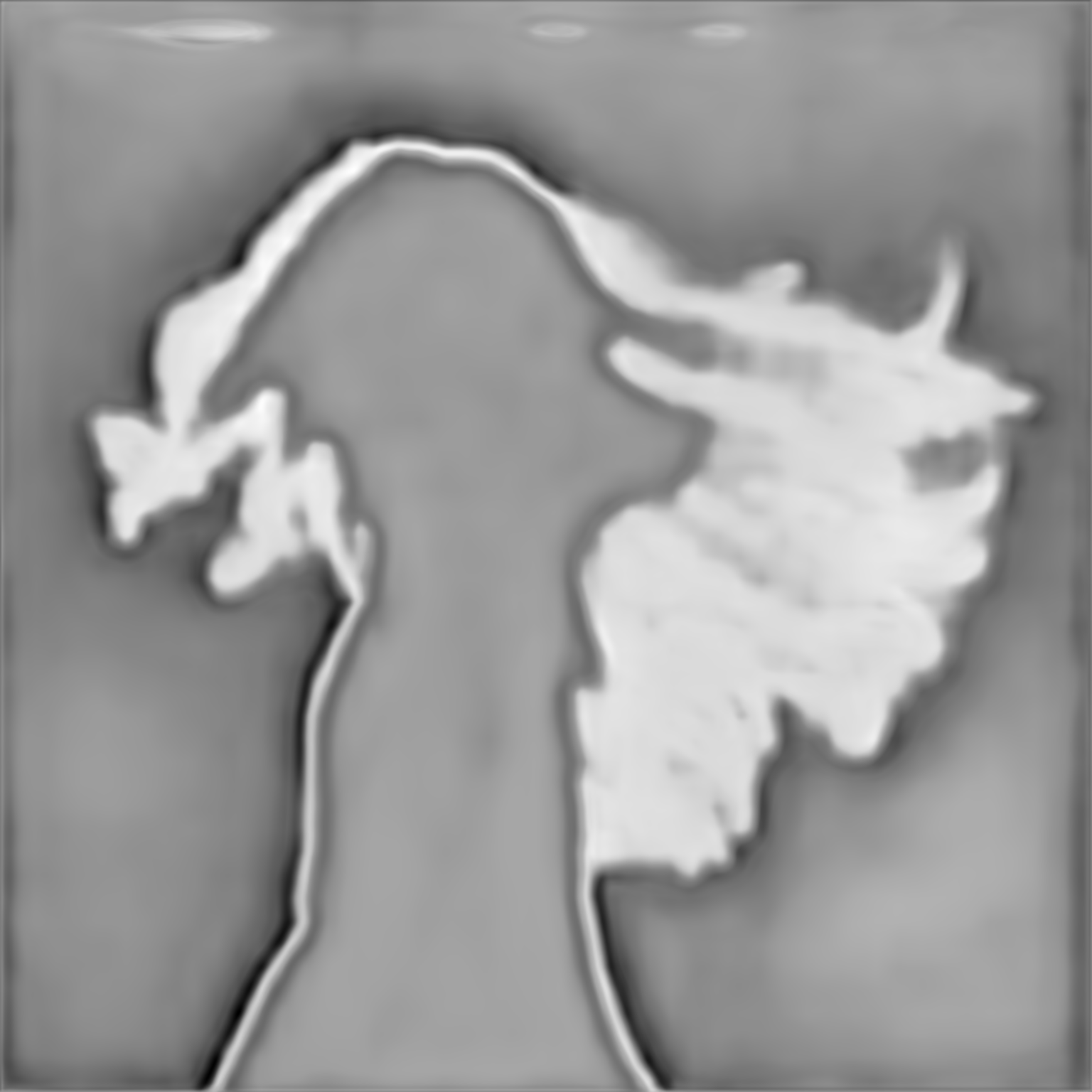}}
        \caption{s3}
        \label{fig:test_4_s3}
    \end{subfigure}
    \centering
    \begin{subfigure}{0.19\linewidth}
        \centerline{\includegraphics[width=0.95\textwidth]{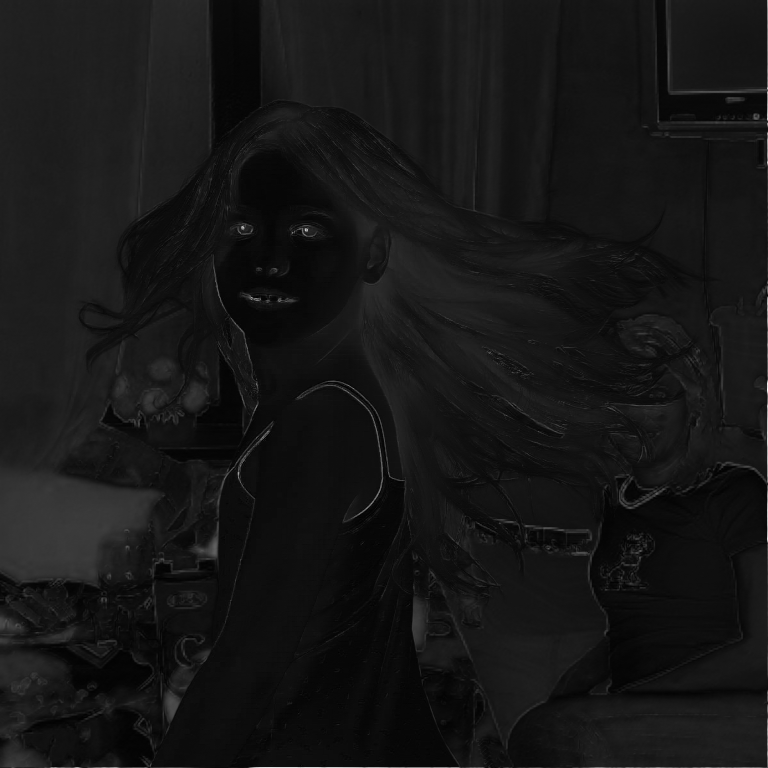}}
        \caption{Low-level feature}
        \label{fig:test_4_low_level_fea}
    \end{subfigure}
    \centering
    \begin{subfigure}{0.19\linewidth}
        \centerline{\includegraphics[width=0.95\textwidth]{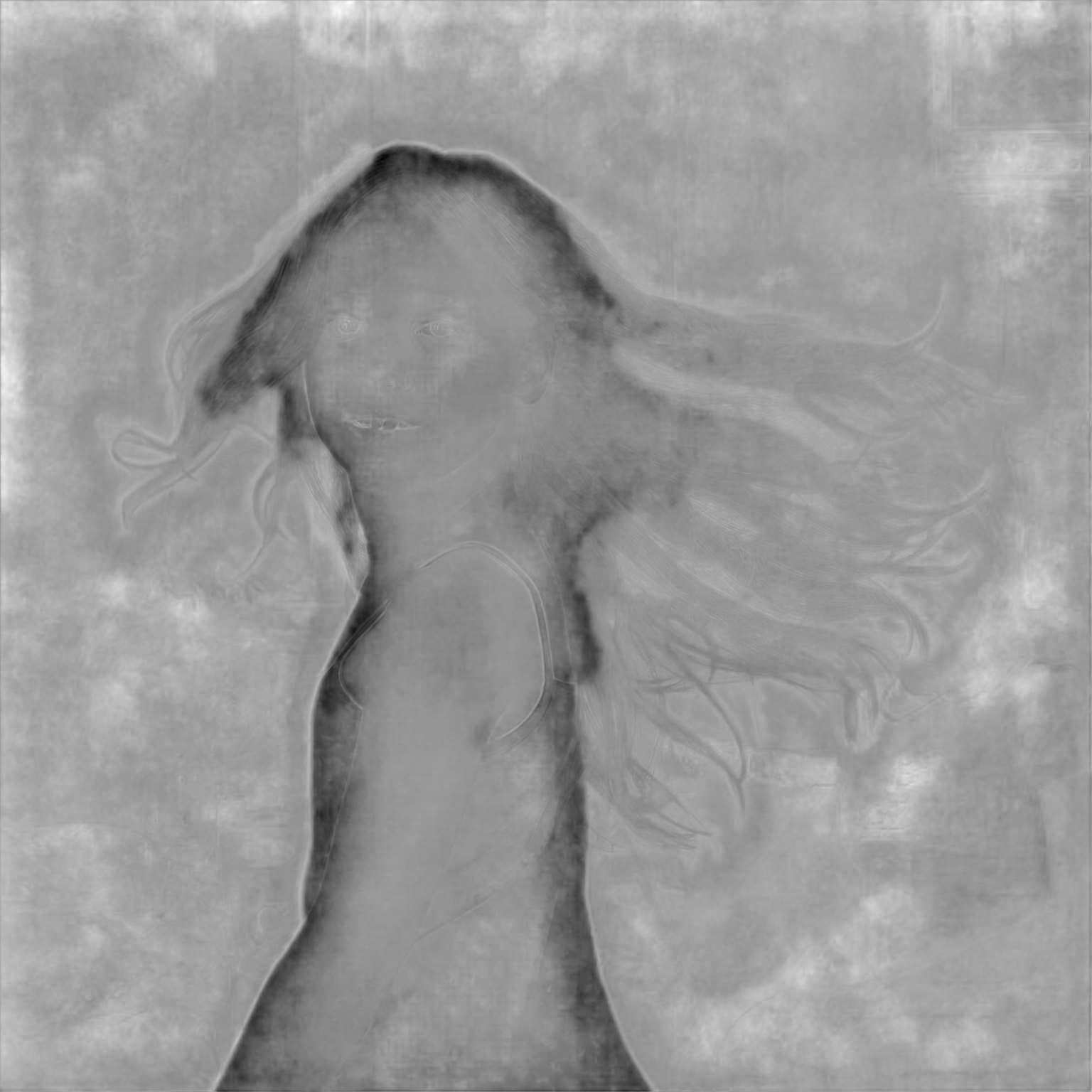}}
        \caption{g1}
        \label{fig:test_4_g1}
    \end{subfigure}
    \centering
    \begin{subfigure}{0.19\linewidth}
        \centerline{\includegraphics[width=0.95\textwidth]{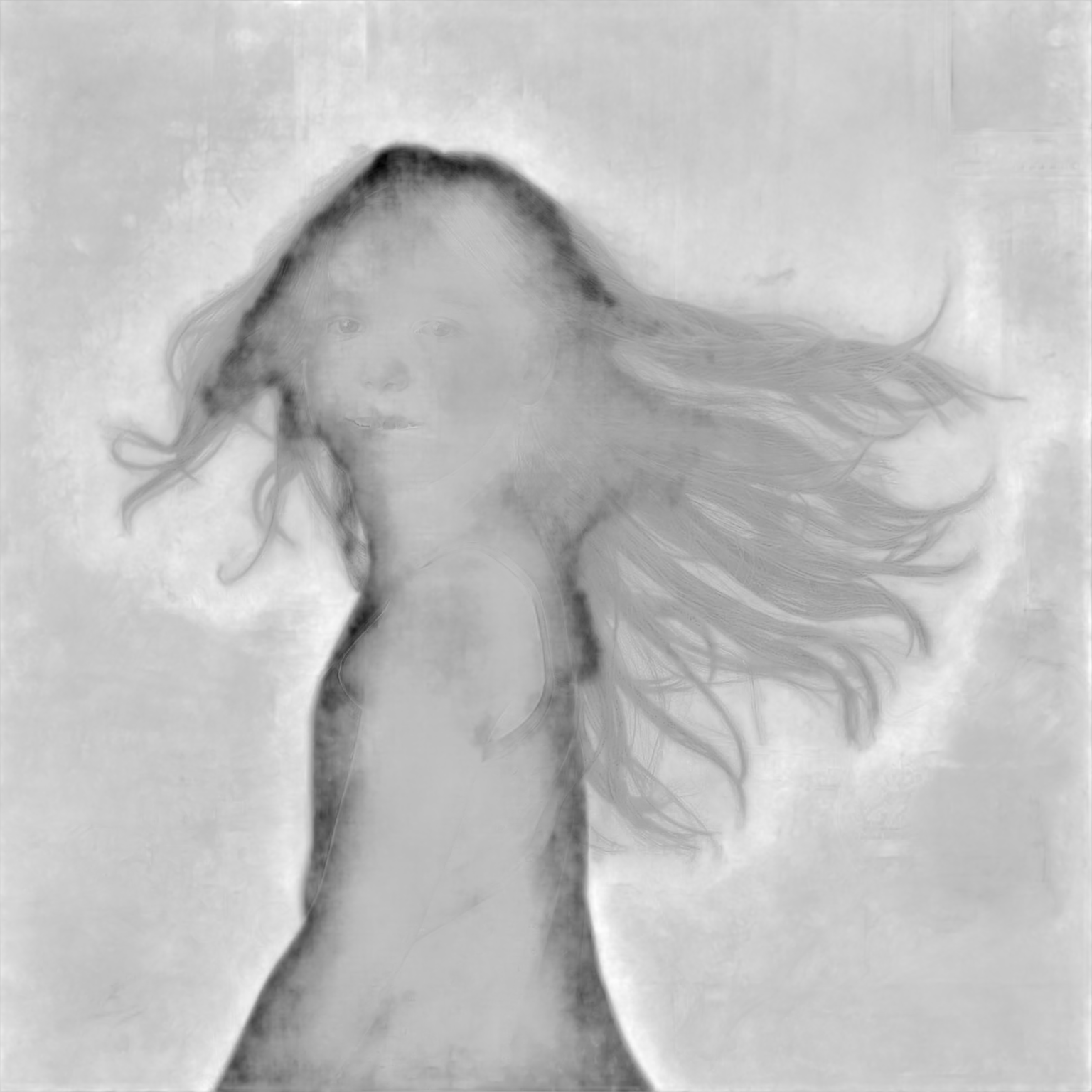}}
        \caption{g2}
        \label{fig:test_4_g2}
    \end{subfigure}
    \centering
    \begin{subfigure}{0.19\linewidth}
        \centerline{\includegraphics[width=0.95\textwidth]{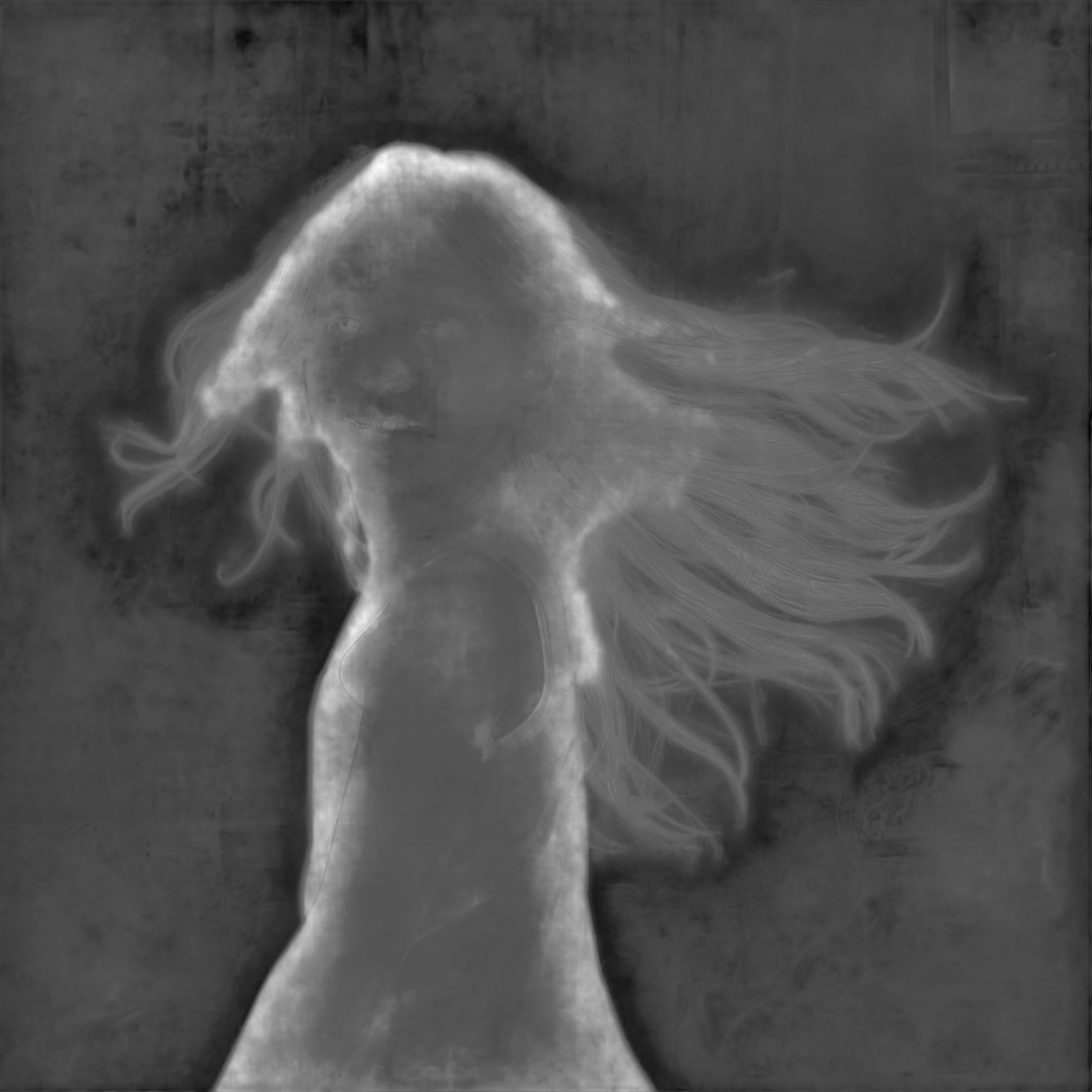}}
        \caption{g3}
        \label{fig:test_4_g3}
    \end{subfigure}
    \centering
    \begin{subfigure}{0.19\linewidth}
        \centerline{\includegraphics[width=0.95\textwidth]{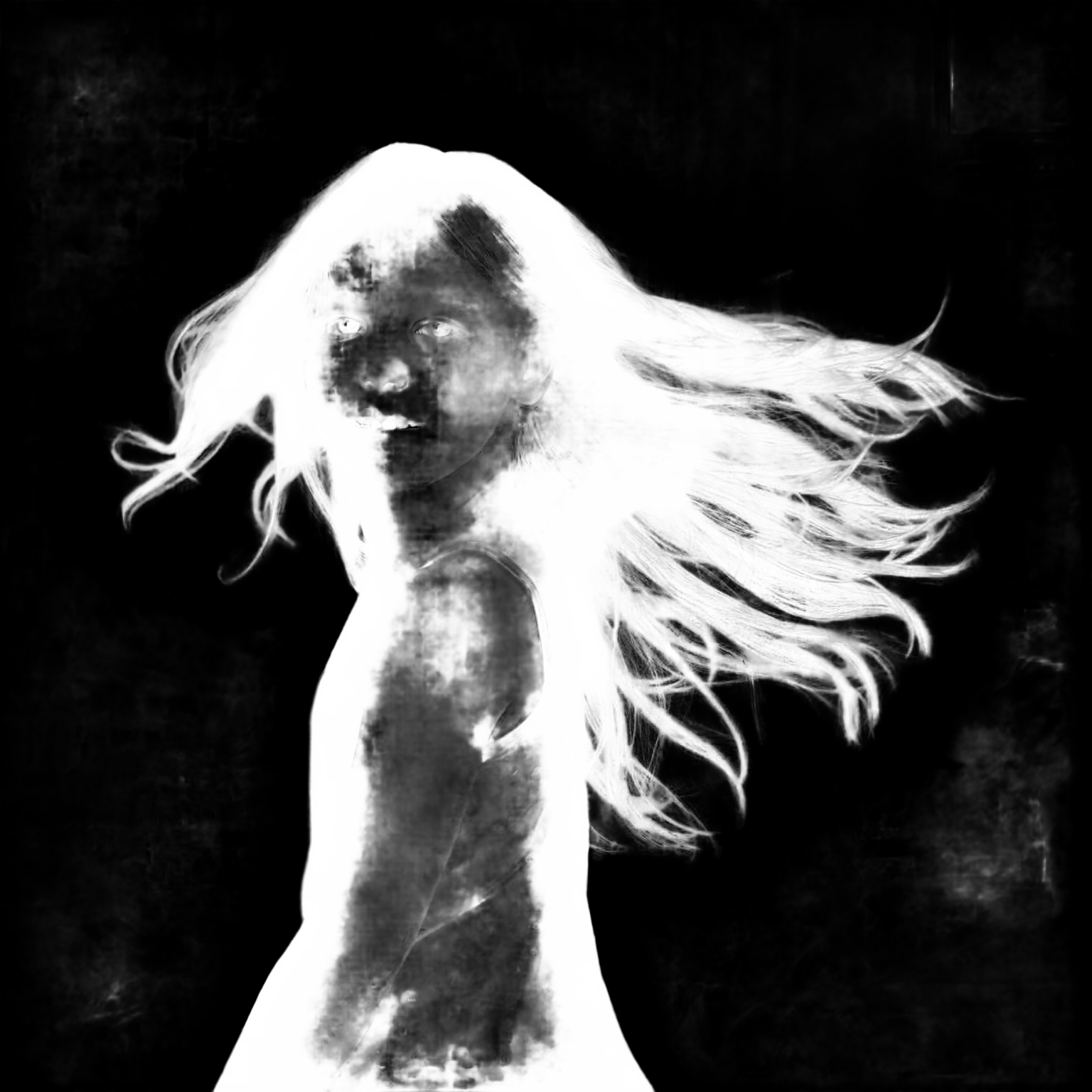}}
        \caption{Detail map}
        \label{fig:test_4_detail}
    \end{subfigure}
\end{figure*}
    
\section{Human Matting}
\begin{figure*}
  \centering
  \begin{subfigure}{0.16\linewidth}
    \centerline{\includegraphics[width=0.95\textwidth]{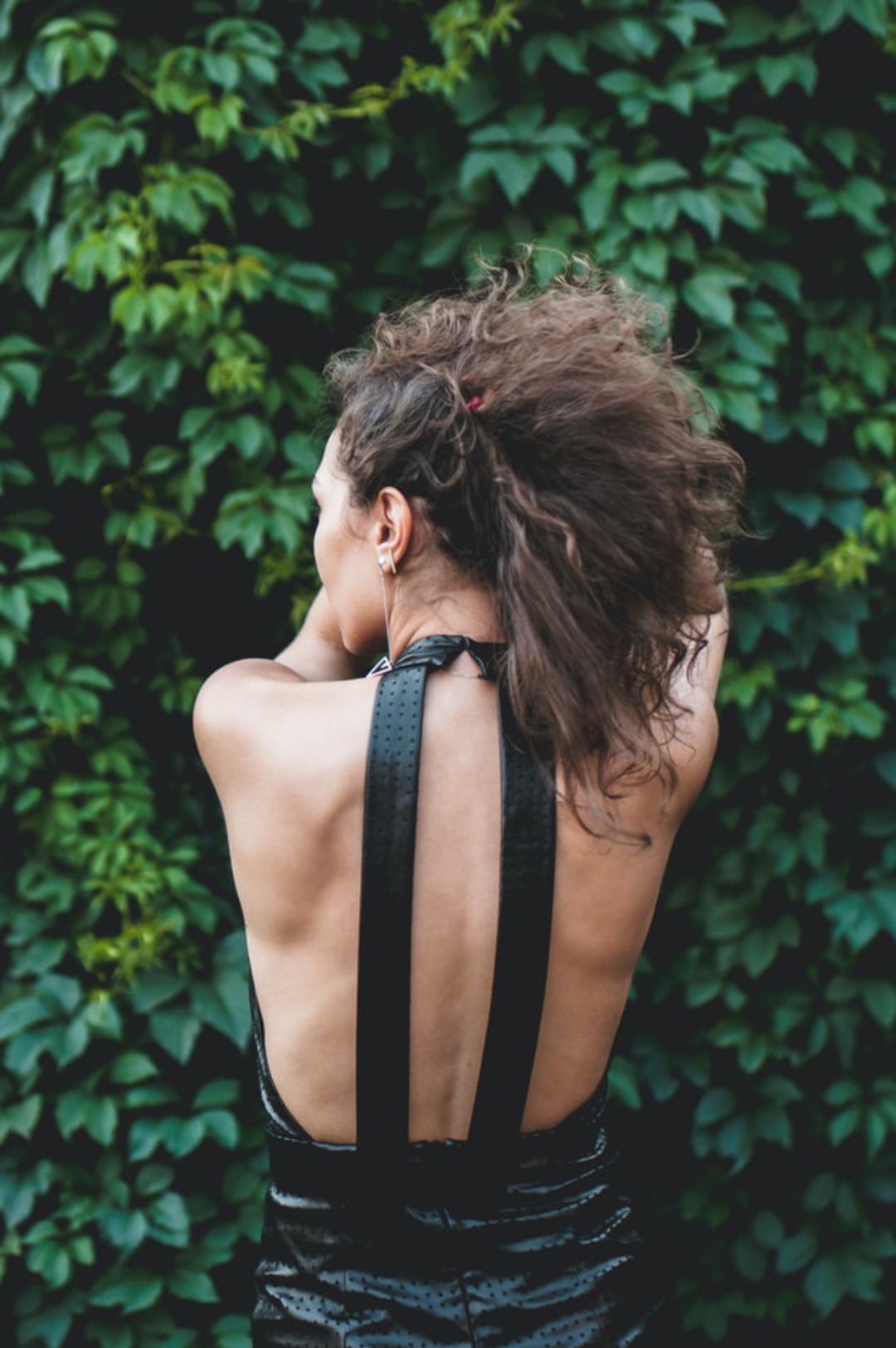}}
    \label{fig:a_image}
  \end{subfigure}
  \centering
  \begin{subfigure}{0.16\linewidth}
    \centerline{\includegraphics[width=0.95\textwidth]{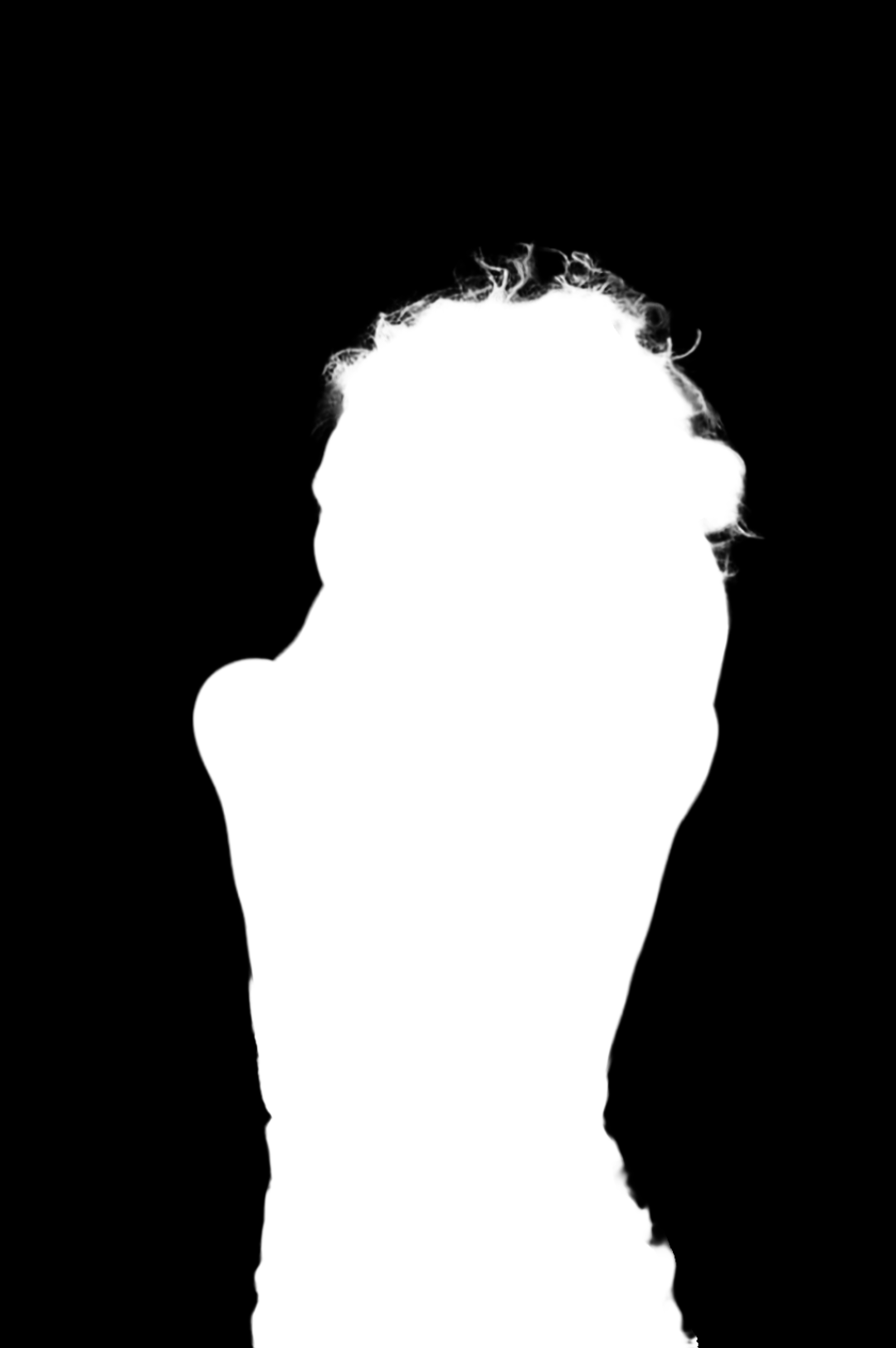}}
    \label{fig:a_alpha}
  \end{subfigure}
    \centering
  \begin{subfigure}{0.16\linewidth}
    \centerline{\includegraphics[width=0.95\textwidth]{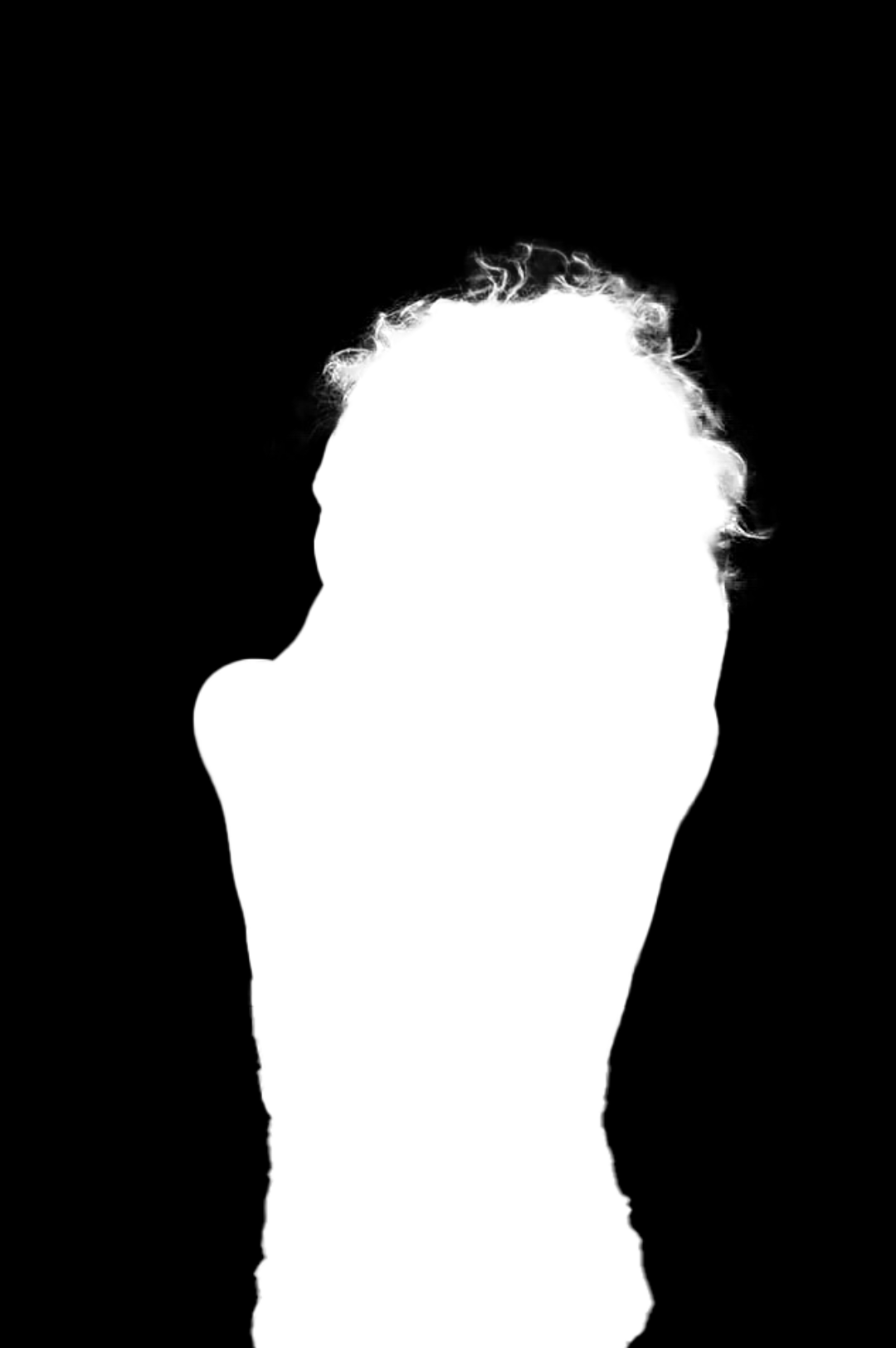}}
    \label{fig:a_gt}
  \end{subfigure}
  \centering
  \begin{subfigure}{0.16\linewidth}
    \centerline{\includegraphics[width=0.95\textwidth]{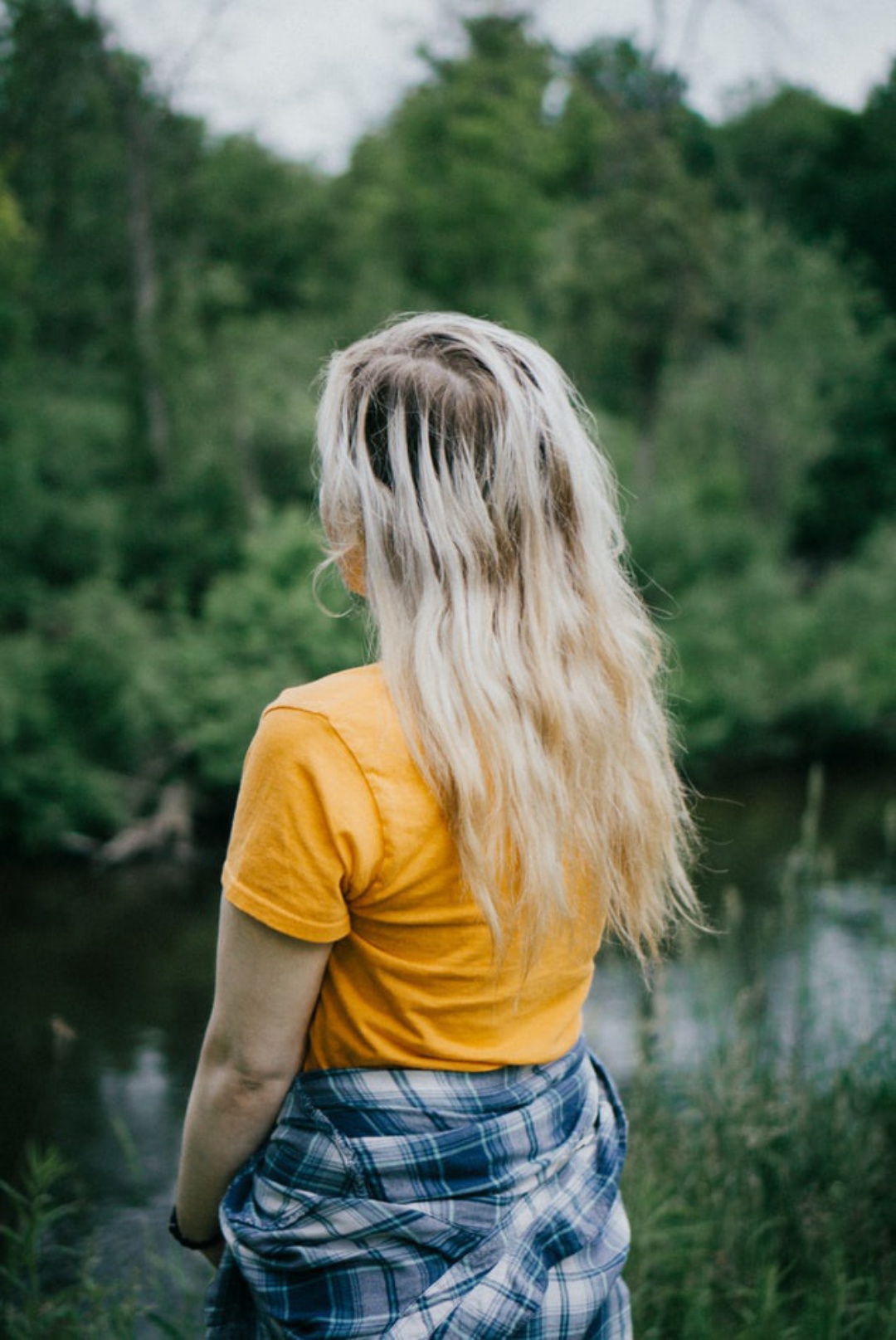}}
    \label{fig:b_image}
  \end{subfigure}
  \centering
  \begin{subfigure}{0.16\linewidth}
    \centerline{\includegraphics[width=0.95\textwidth]{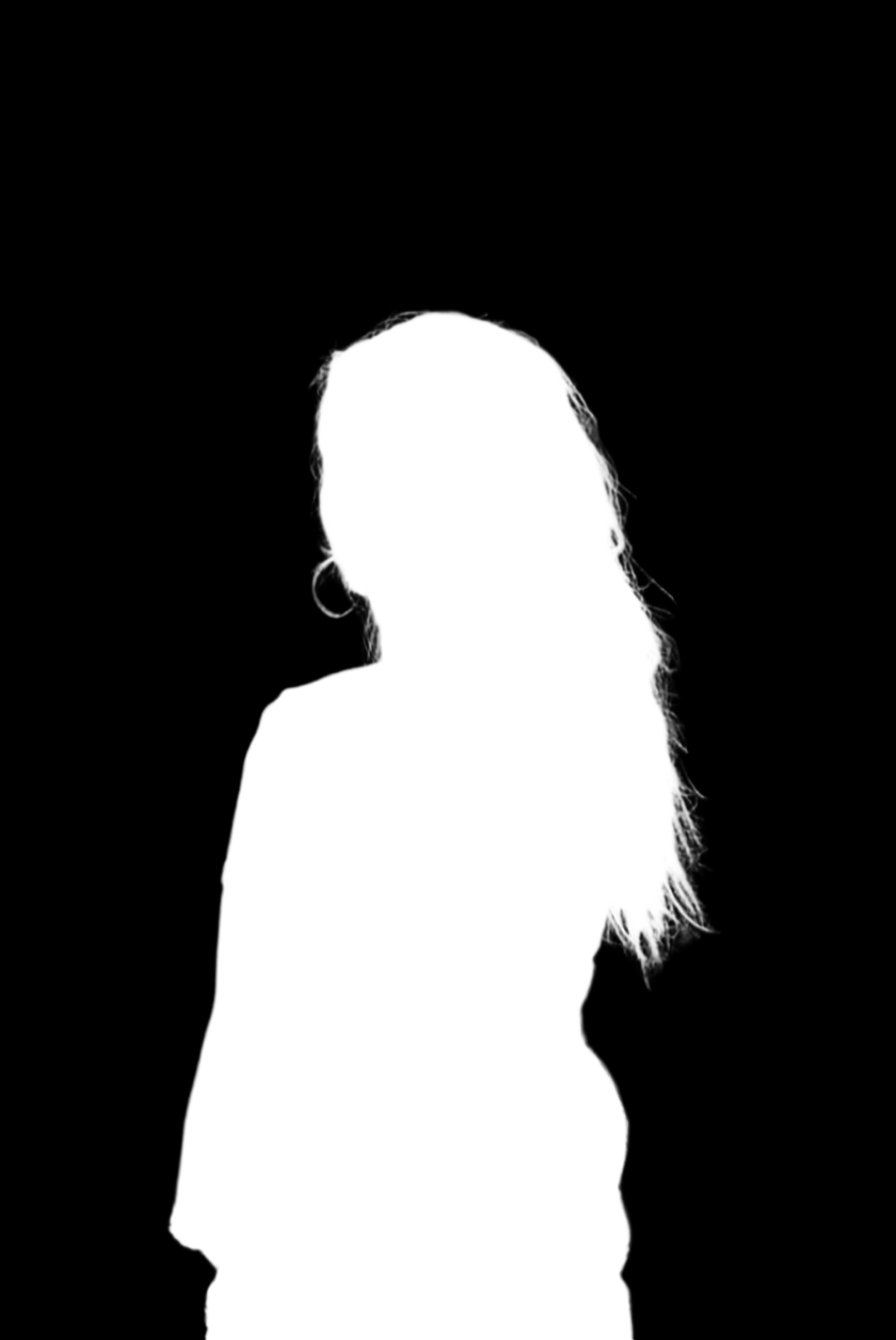}}
    \label{fig:b_alpha}
  \end{subfigure}
  \centering
  \begin{subfigure}{0.16\linewidth}
    \centerline{\includegraphics[width=0.95\textwidth]{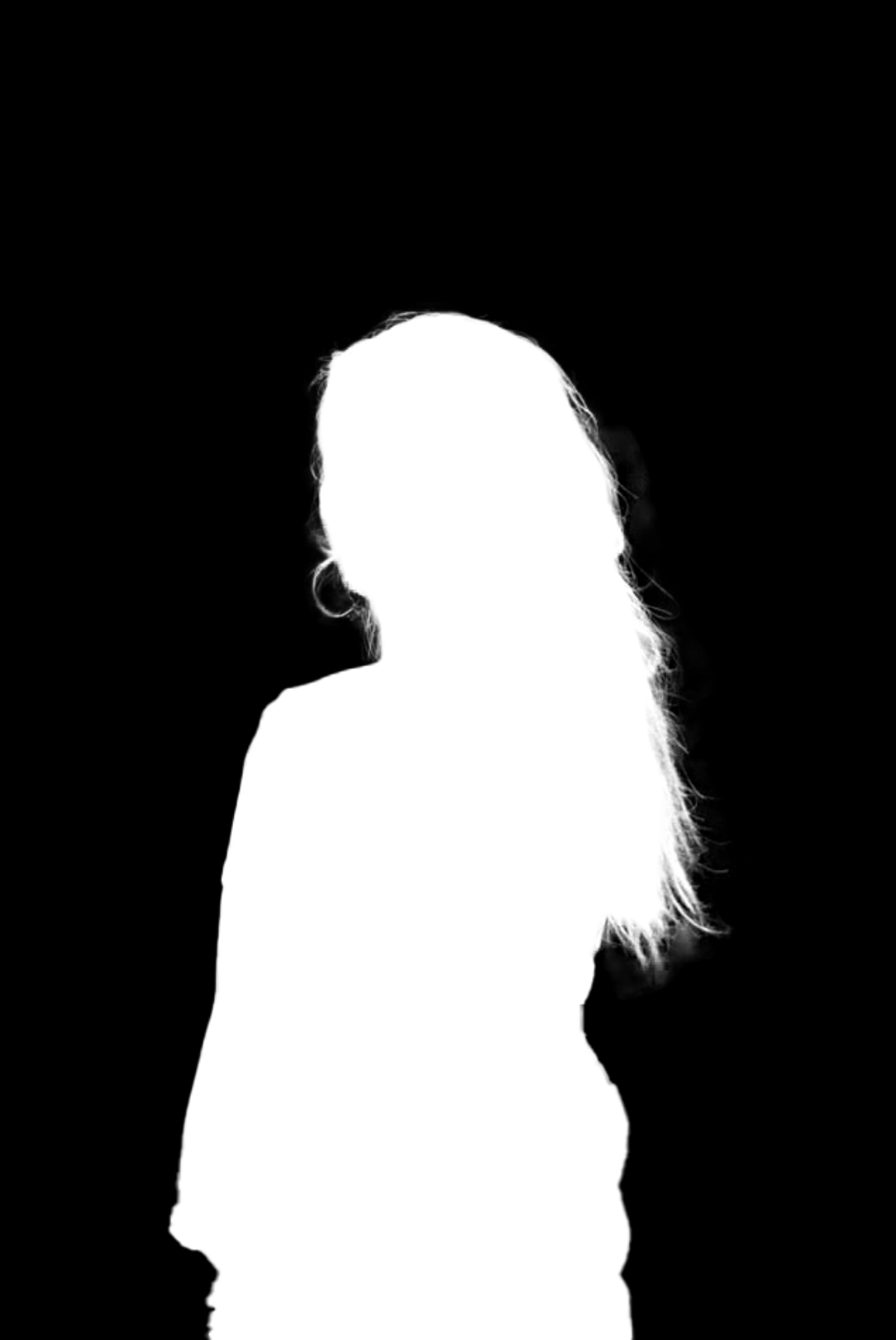}}
    \label{fig:b_gt}
  \end{subfigure}
  \centering
  \begin{subfigure}{0.16\linewidth}
    \centerline{\includegraphics[width=0.95\textwidth]{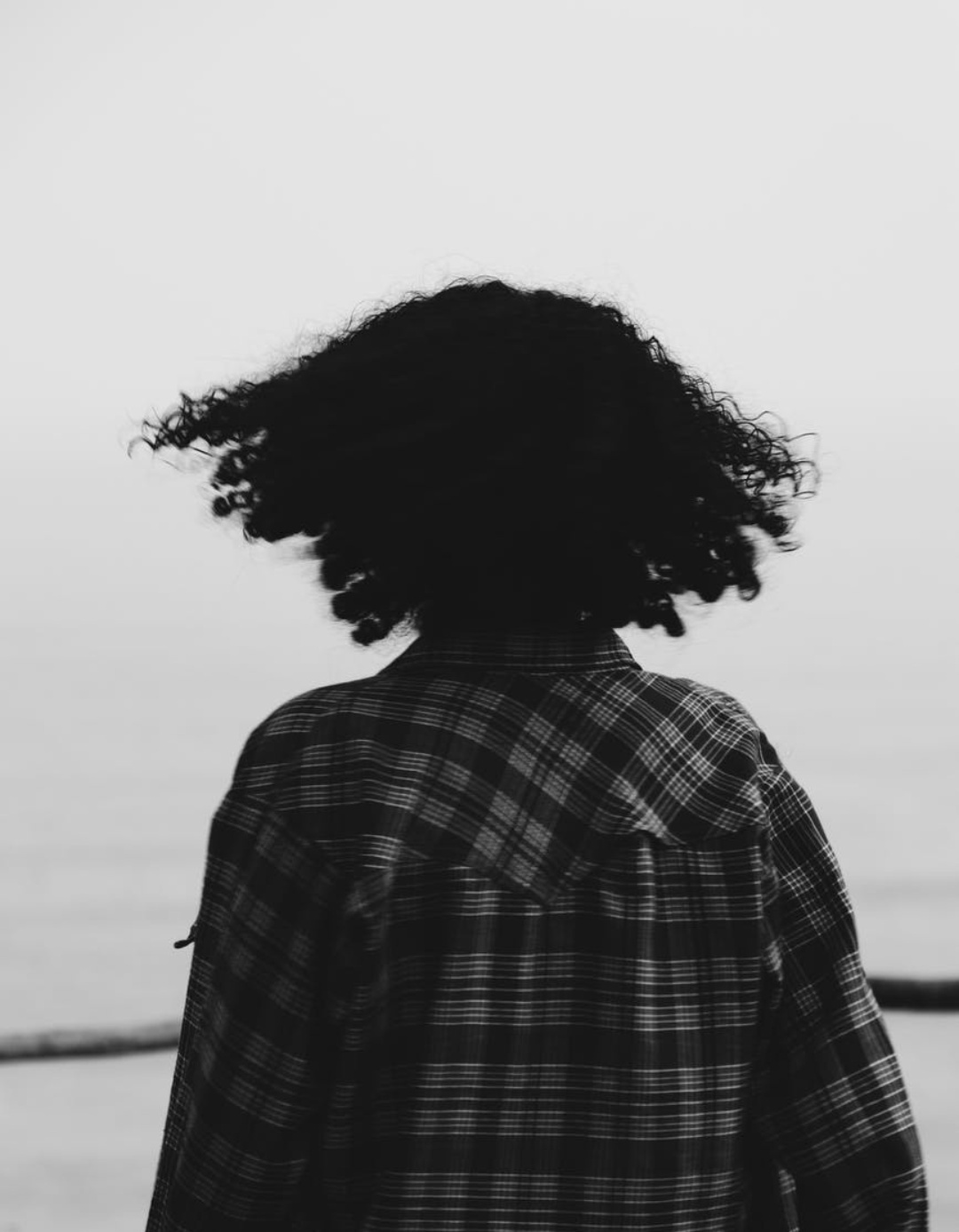}}
    \label{fig:c_image}
  \end{subfigure}
  \centering
  \begin{subfigure}{0.16\linewidth}
    \centerline{\includegraphics[width=0.95\textwidth]{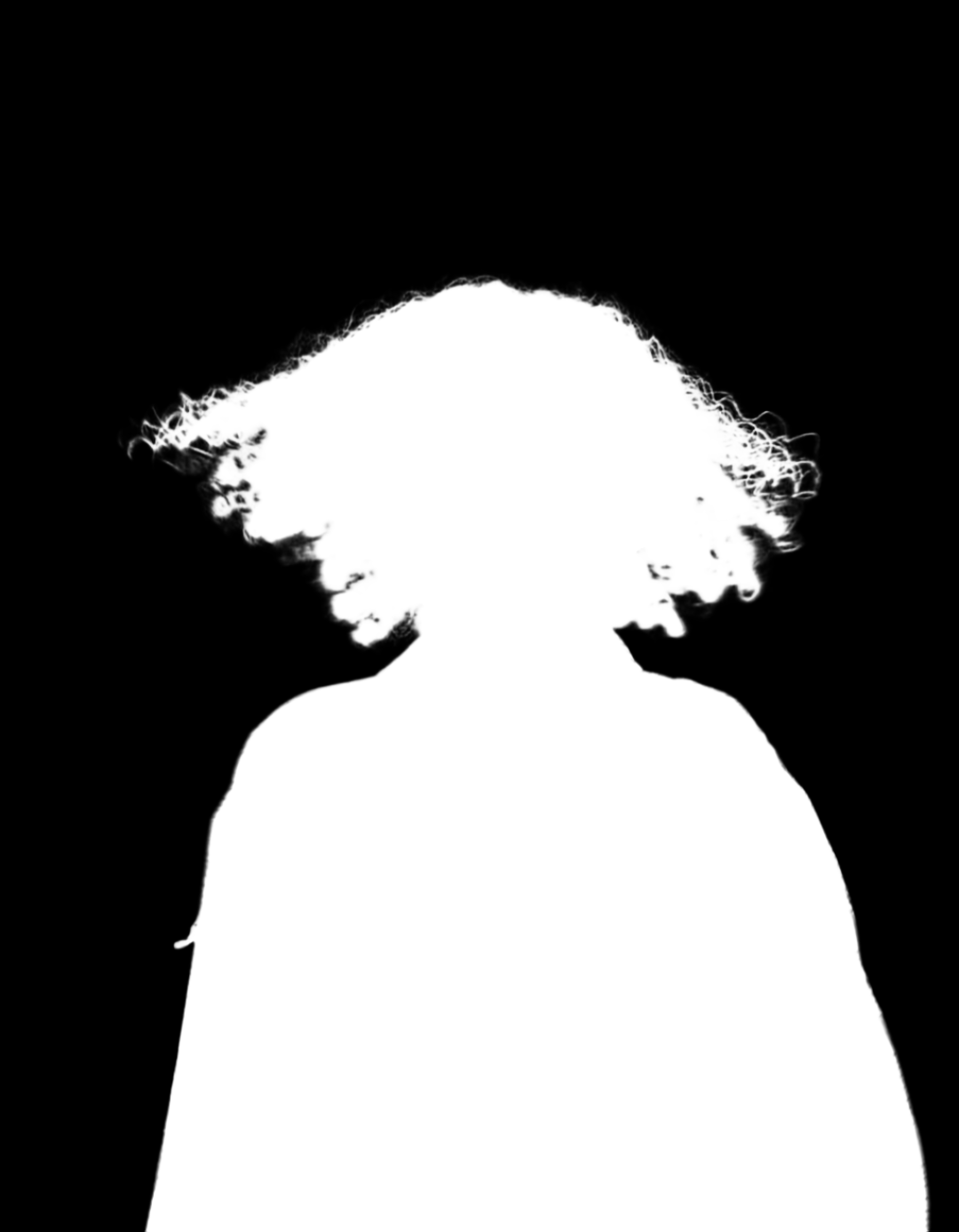}}
    \label{fig:c_alpha}
  \end{subfigure}
  \centering
  \begin{subfigure}{0.16\linewidth}
    \centerline{\includegraphics[width=0.95\textwidth]{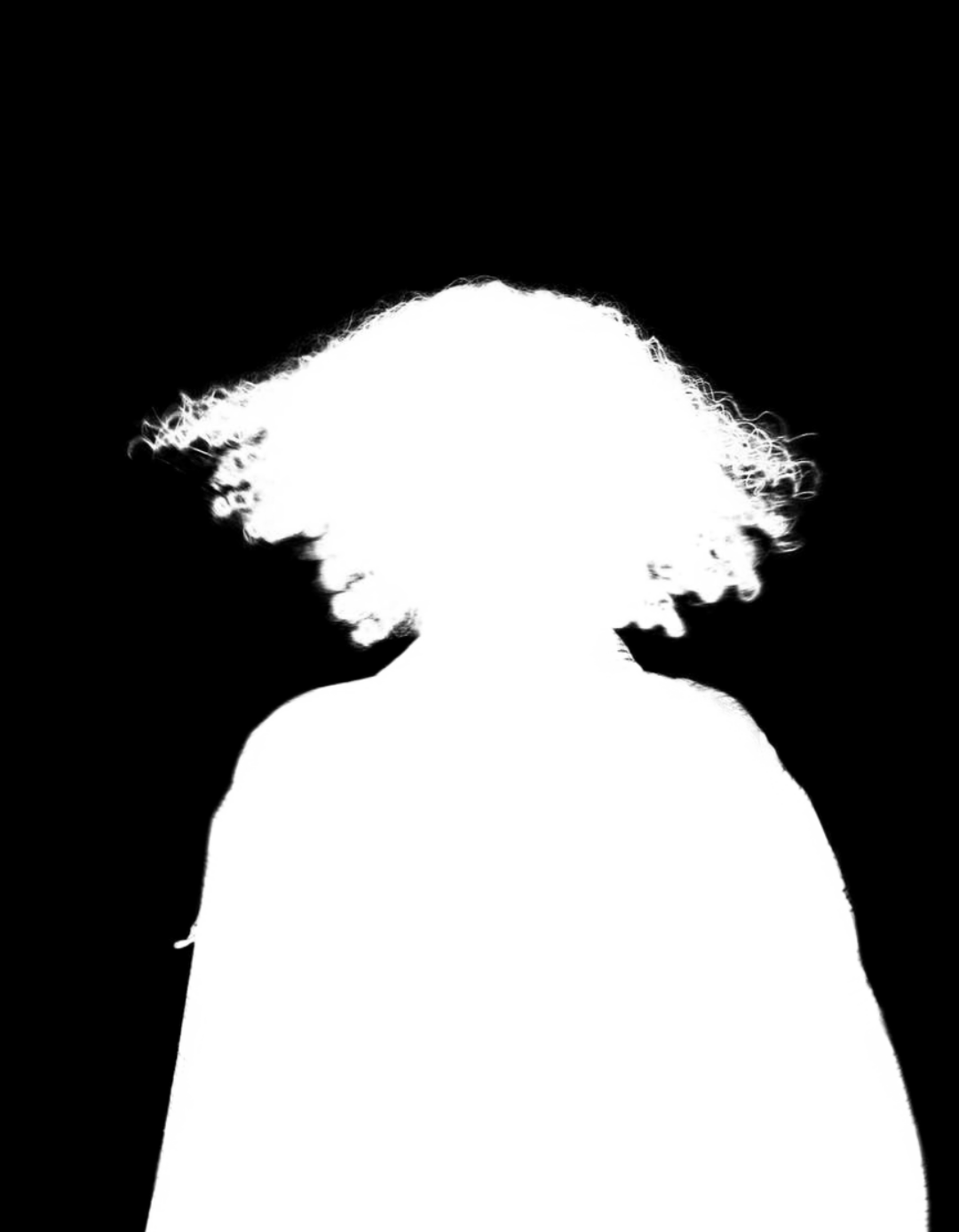}}
    \label{fig:c_gt}
  \end{subfigure}
\centering
  \begin{subfigure}{0.16\linewidth}
    \centerline{\includegraphics[width=0.95\textwidth]{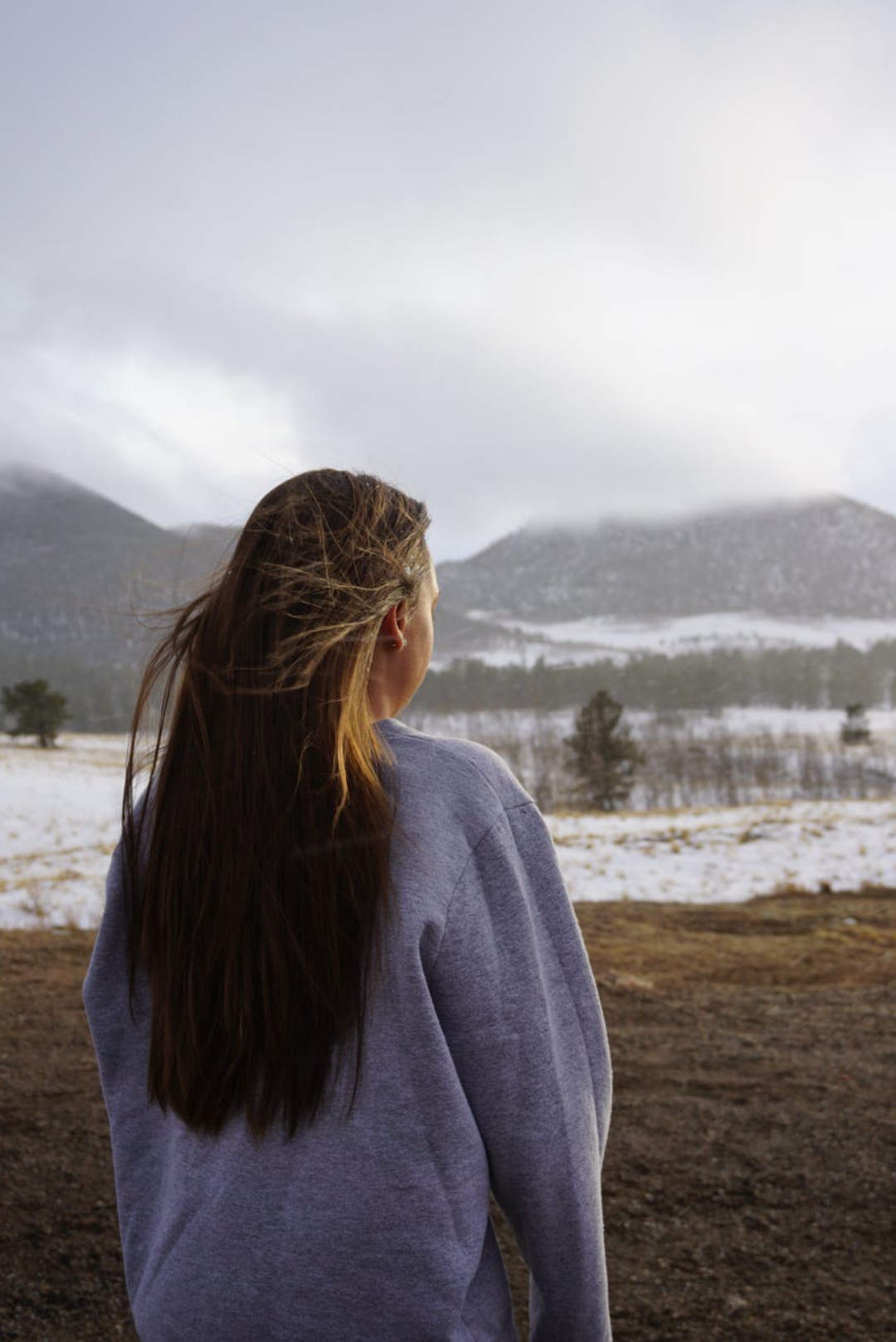}}
    \label{fig:f_image}
  \end{subfigure}
  \centering
  \begin{subfigure}{0.16\linewidth}
    \centerline{\includegraphics[width=0.95\textwidth]{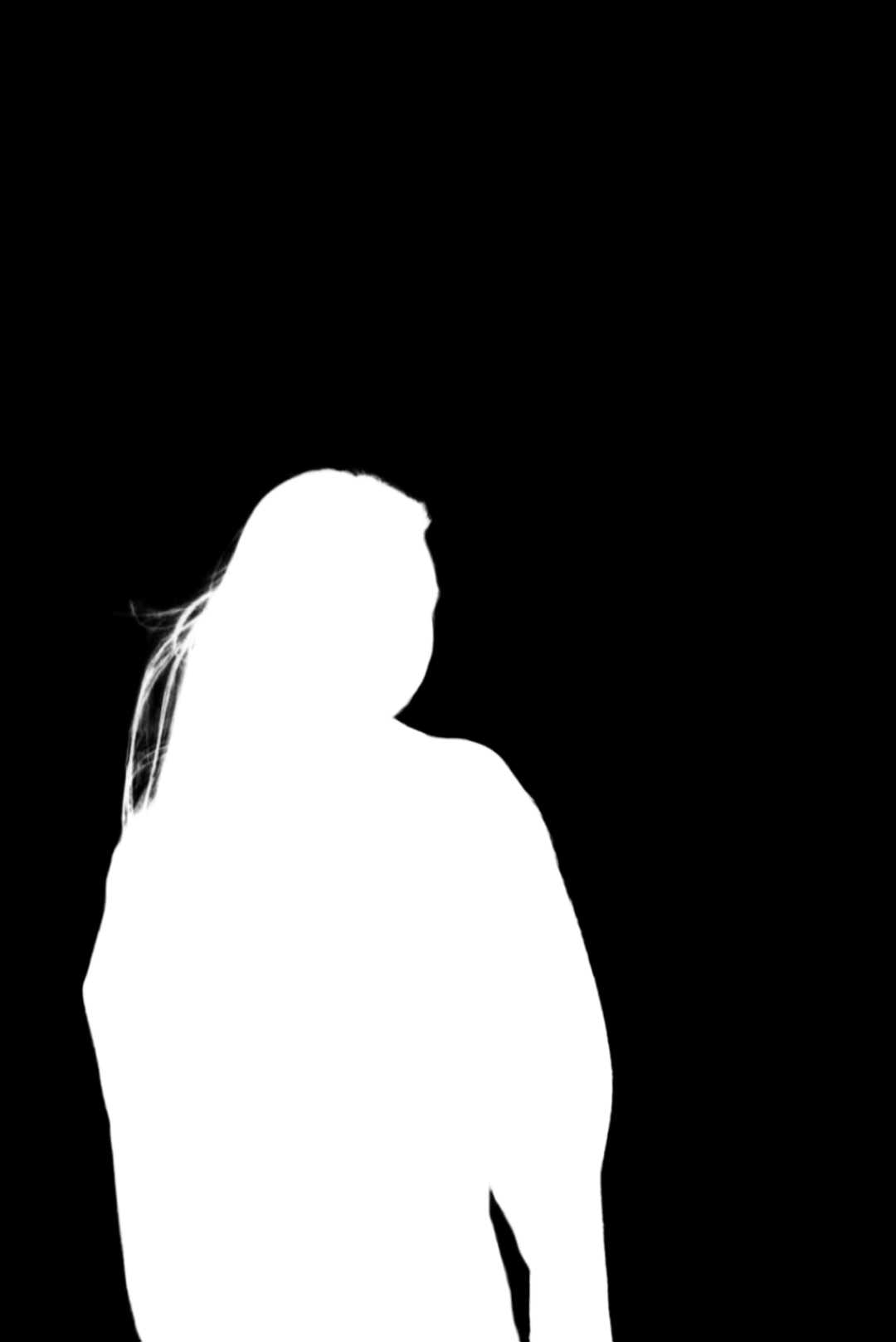}}
    \label{fig:f_alpha}
  \end{subfigure}
  \centering
  \begin{subfigure}{0.16\linewidth}
    \centerline{\includegraphics[width=0.95\textwidth]{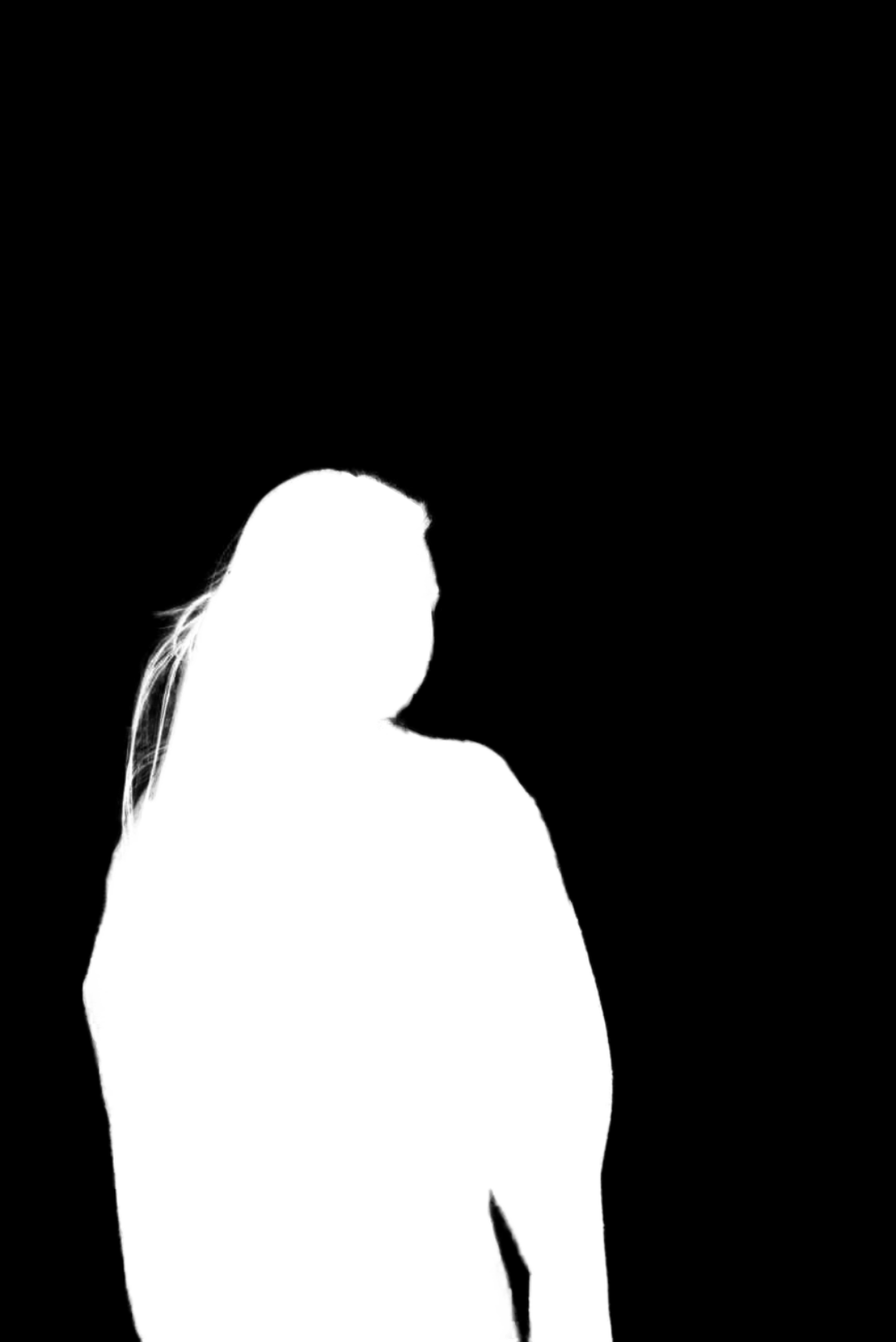}}
    \label{fig:f_gt}
  \end{subfigure}
\centering
  \begin{subfigure}{0.16\linewidth}
    \centerline{\includegraphics[width=0.95\textwidth]{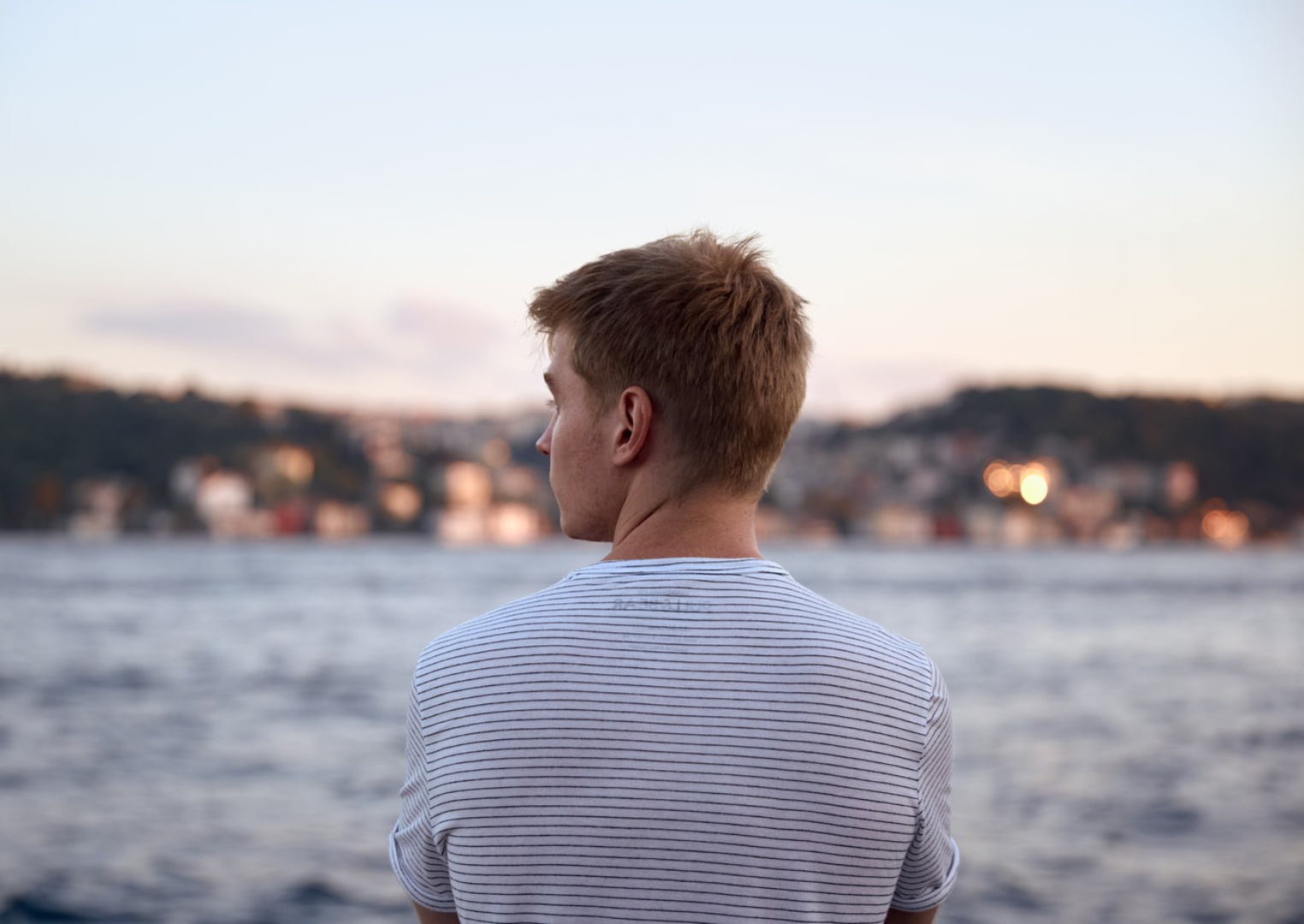}}
    \label{fig:e_image}
  \end{subfigure}
  \centering
  \begin{subfigure}{0.16\linewidth}
    \centerline{\includegraphics[width=0.95\textwidth]{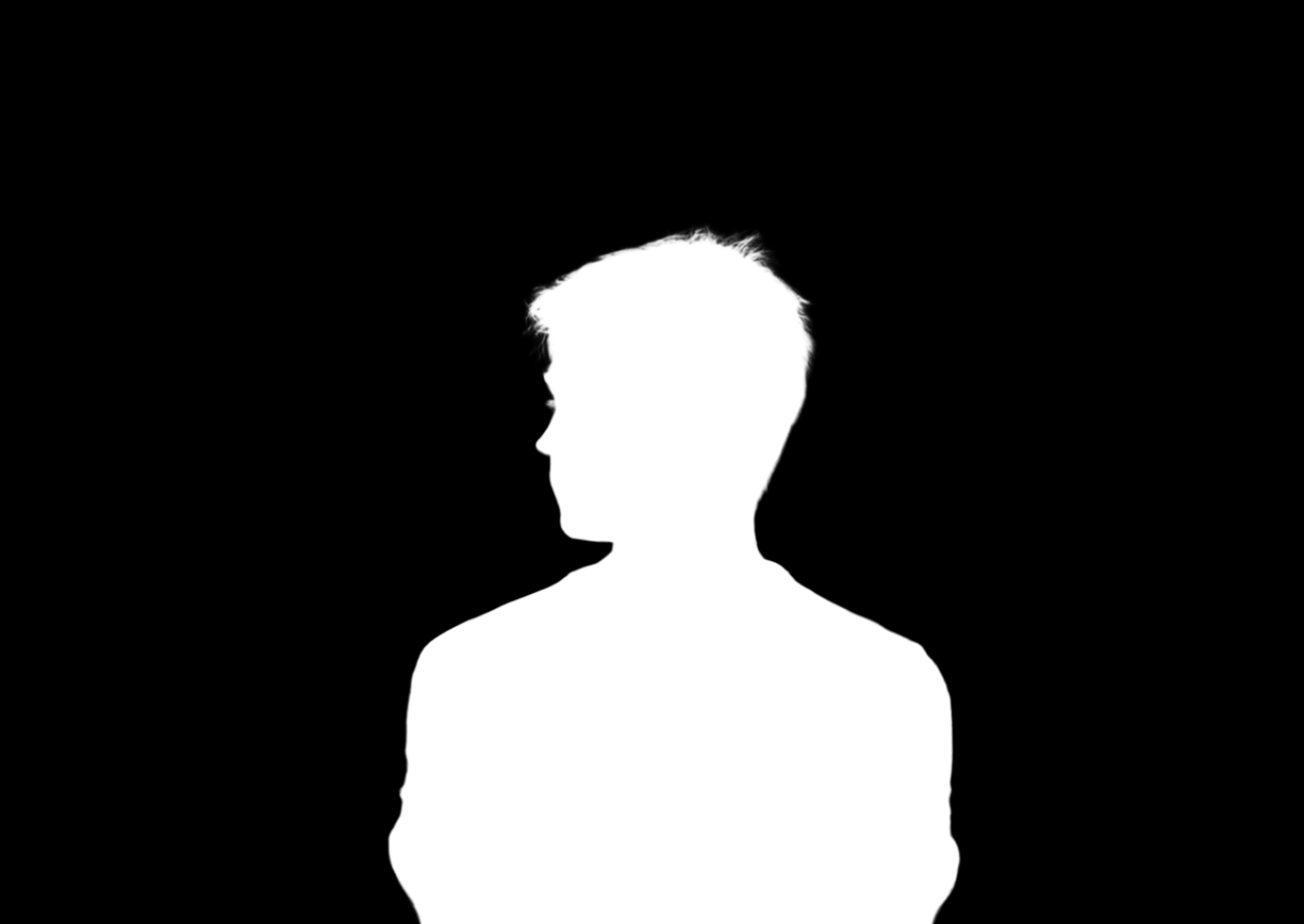}}
    \label{fig:e_alpha}
  \end{subfigure}
  \centering
  \begin{subfigure}{0.16\linewidth}
    \centerline{\includegraphics[width=0.95\textwidth]{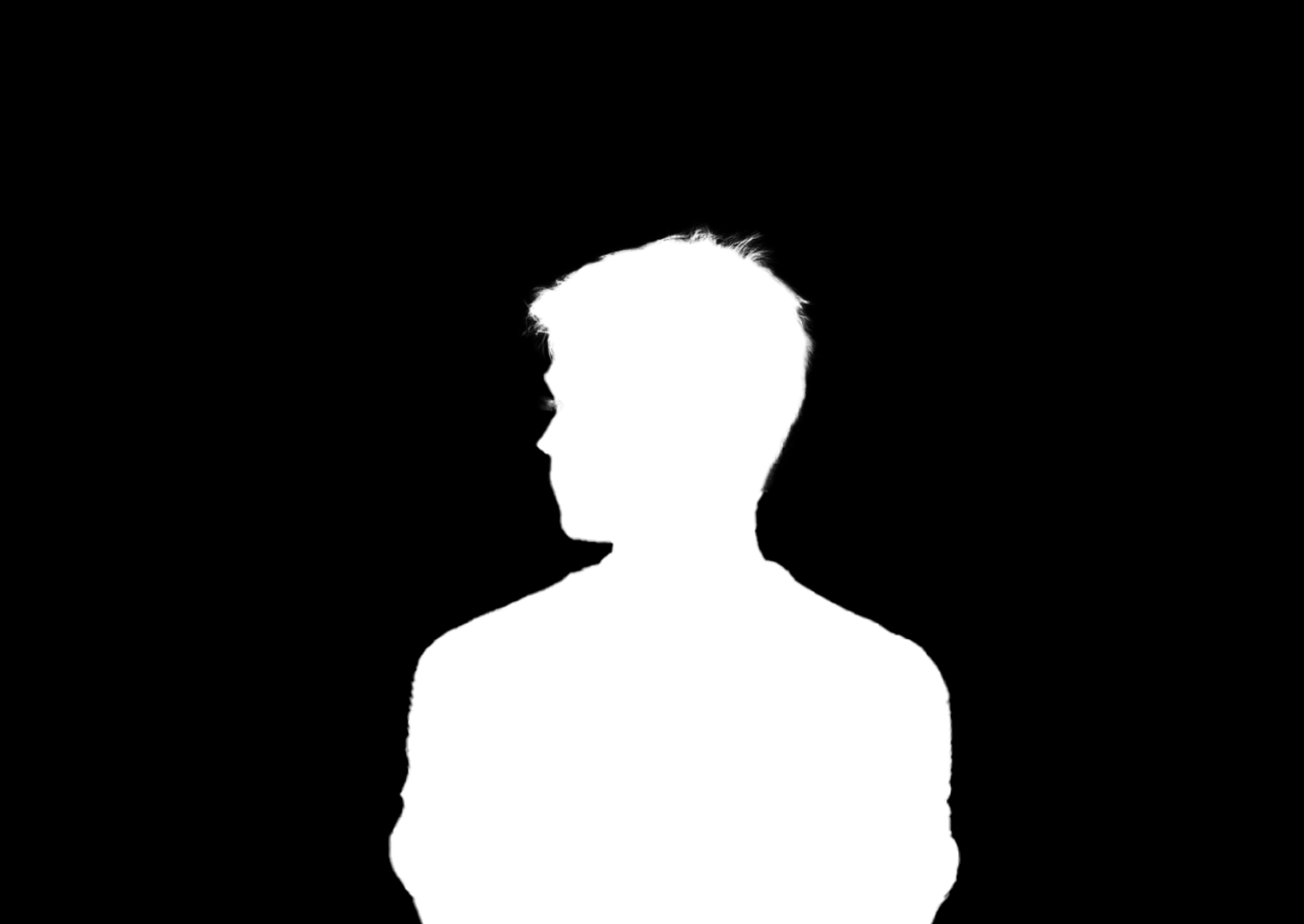}}
    \label{fig:e_gt}
  \end{subfigure}
\centering
  \begin{subfigure}{0.16\linewidth}
    \centerline{\includegraphics[width=0.95\textwidth]{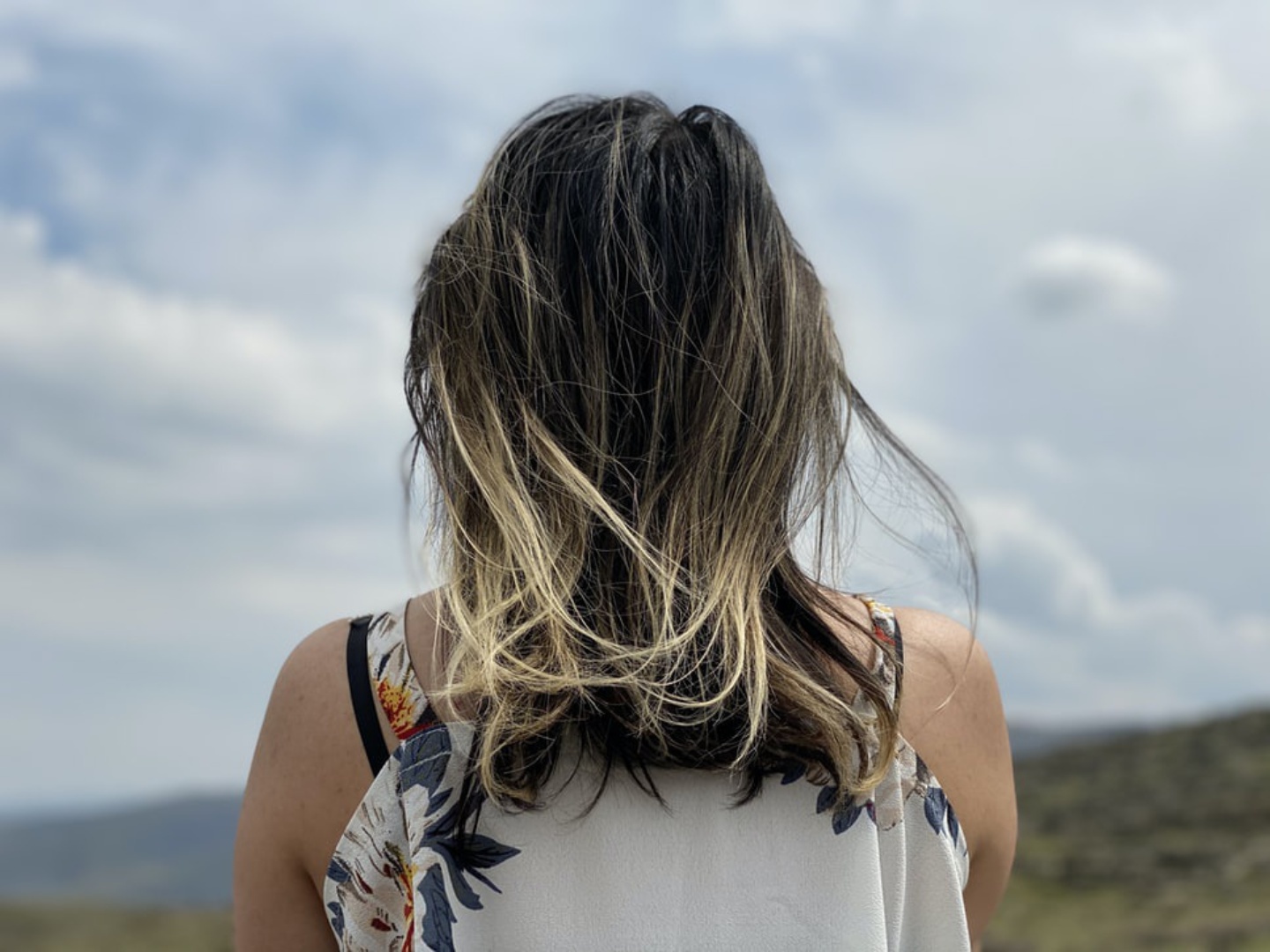}}
    \label{fig:d_image}
  \end{subfigure}
  \centering
  \begin{subfigure}{0.16\linewidth}
    \centerline{\includegraphics[width=0.95\textwidth]{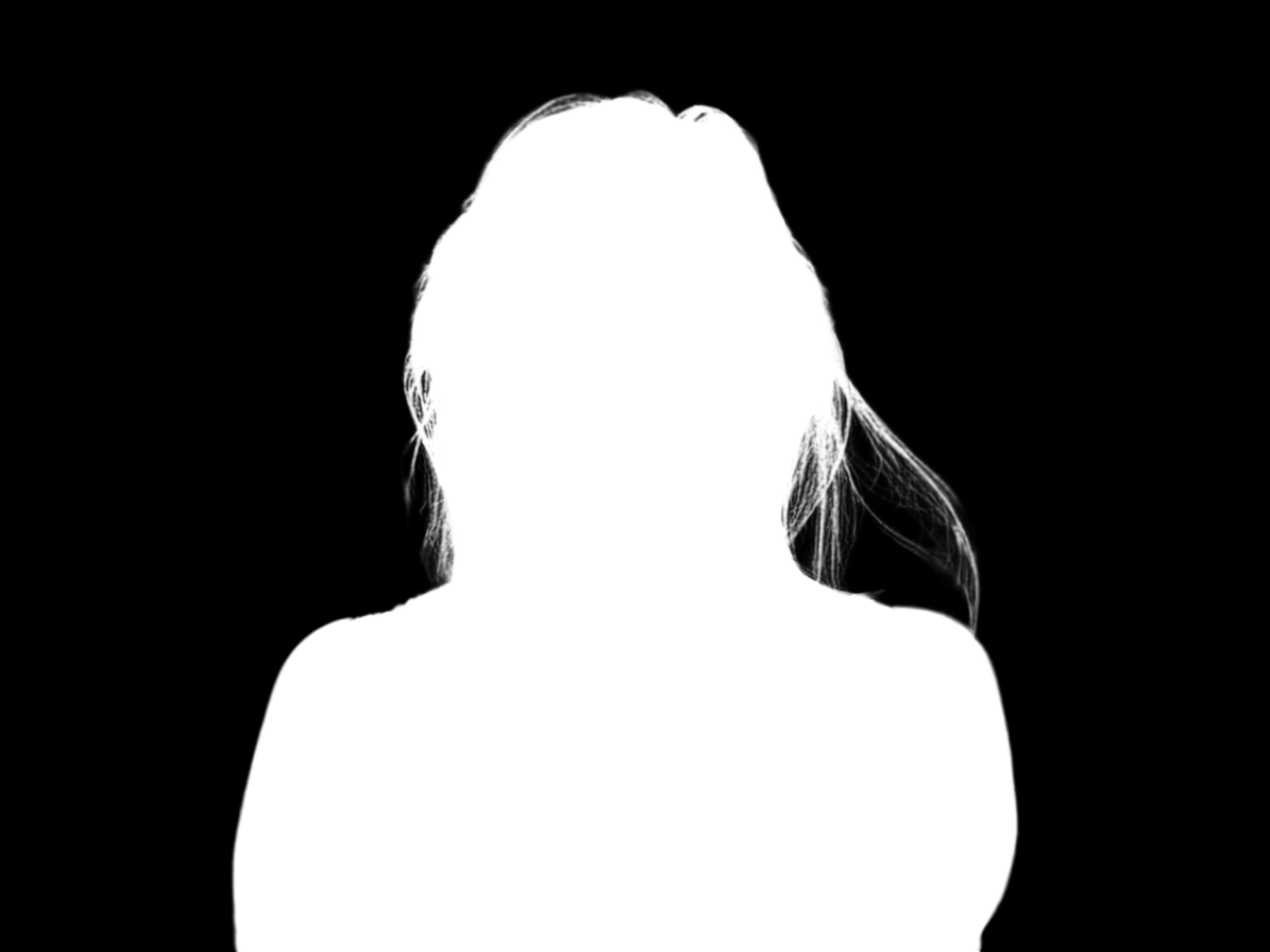}}
    \label{fig:d_alpha}
  \end{subfigure}
  \centering
  \begin{subfigure}{0.16\linewidth}
    \centerline{\includegraphics[width=0.95\textwidth]{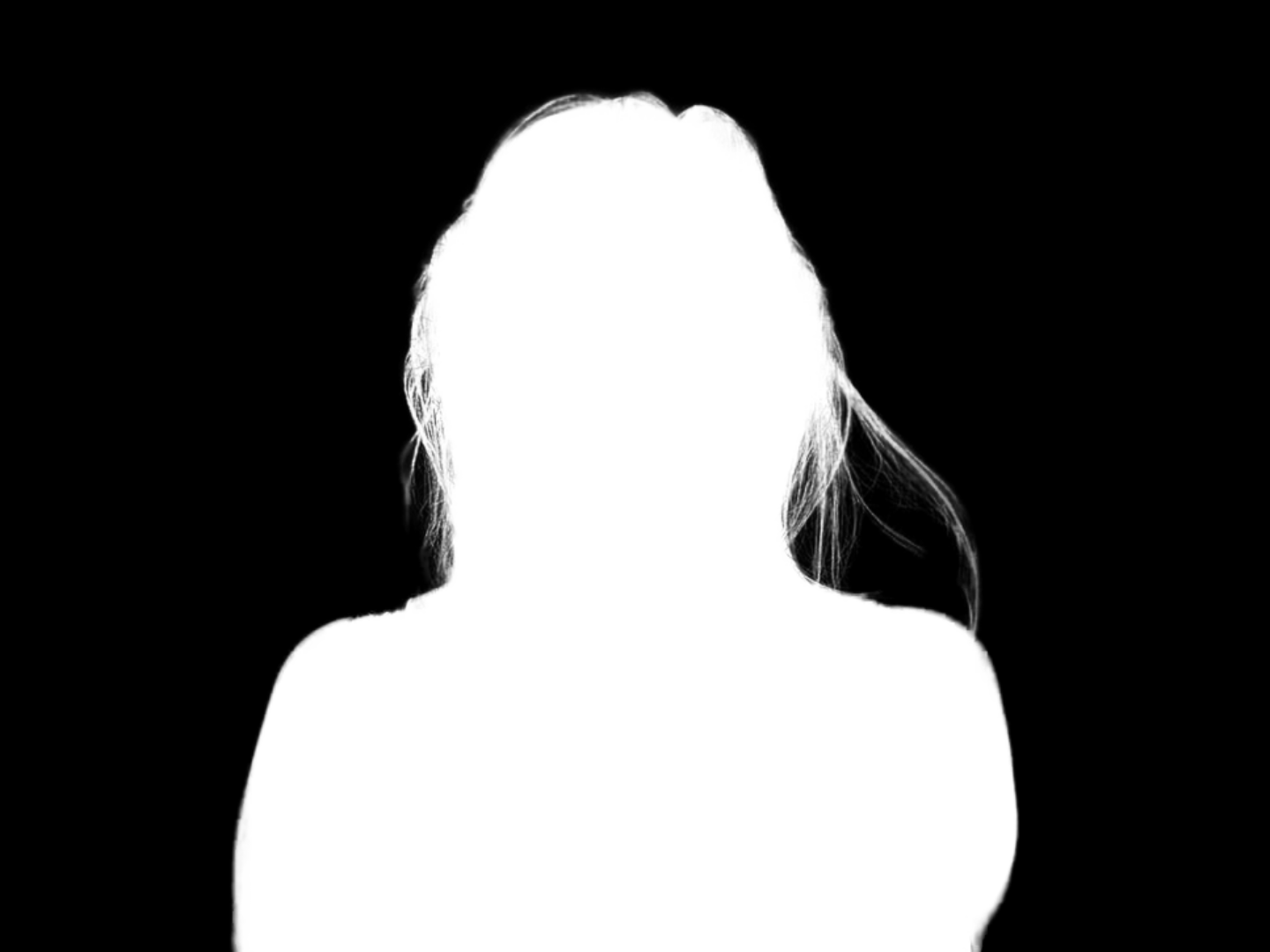}}
    \label{fig:d_gt}
  \end{subfigure}

\centering
  \begin{subfigure}{0.16\linewidth}
    \centerline{\includegraphics[width=0.95\textwidth]{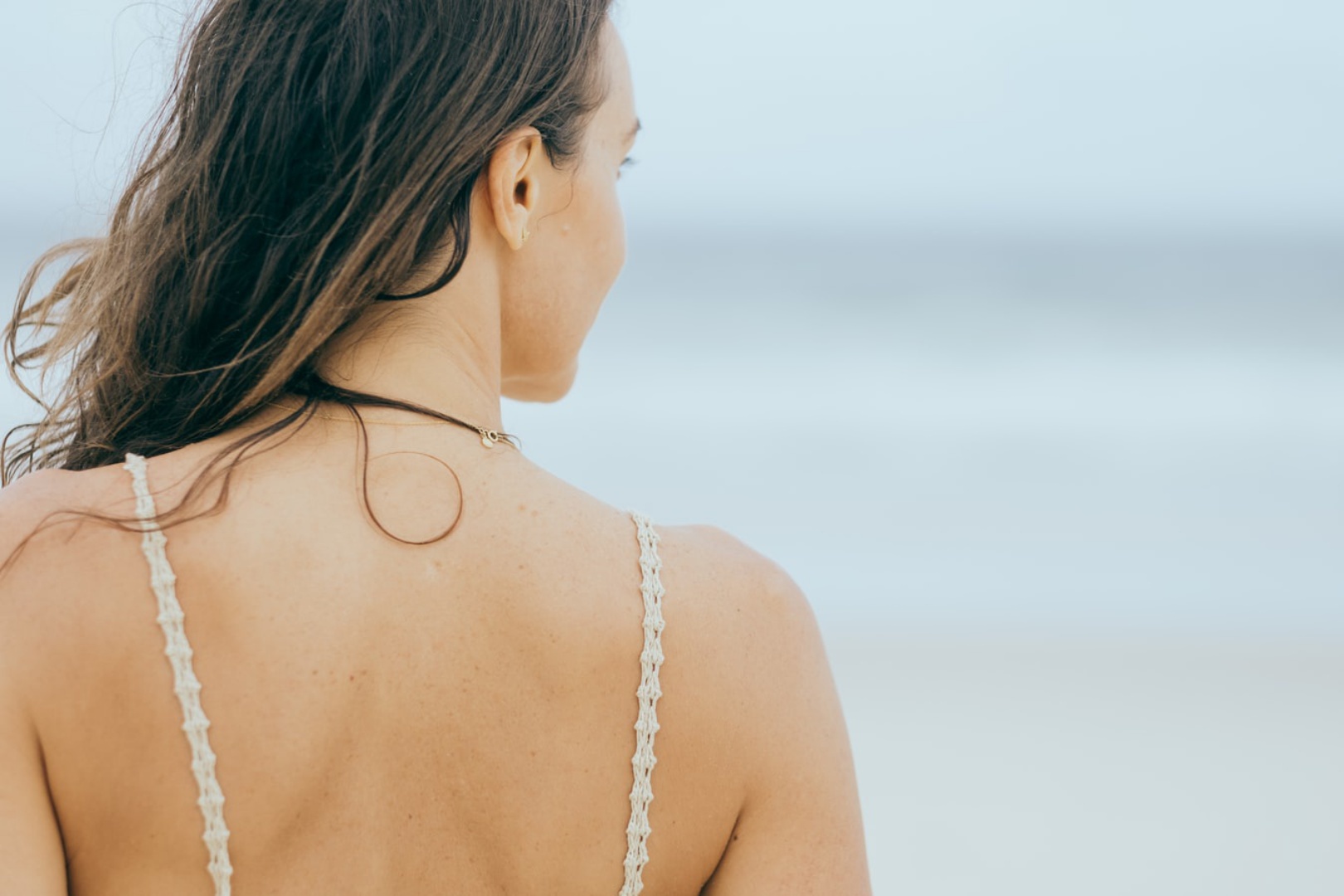}}
    \caption{Image}
    \label{fig:g_image}
  \end{subfigure}
  \centering
  \begin{subfigure}{0.16\linewidth}
    \centerline{\includegraphics[width=0.95\textwidth]{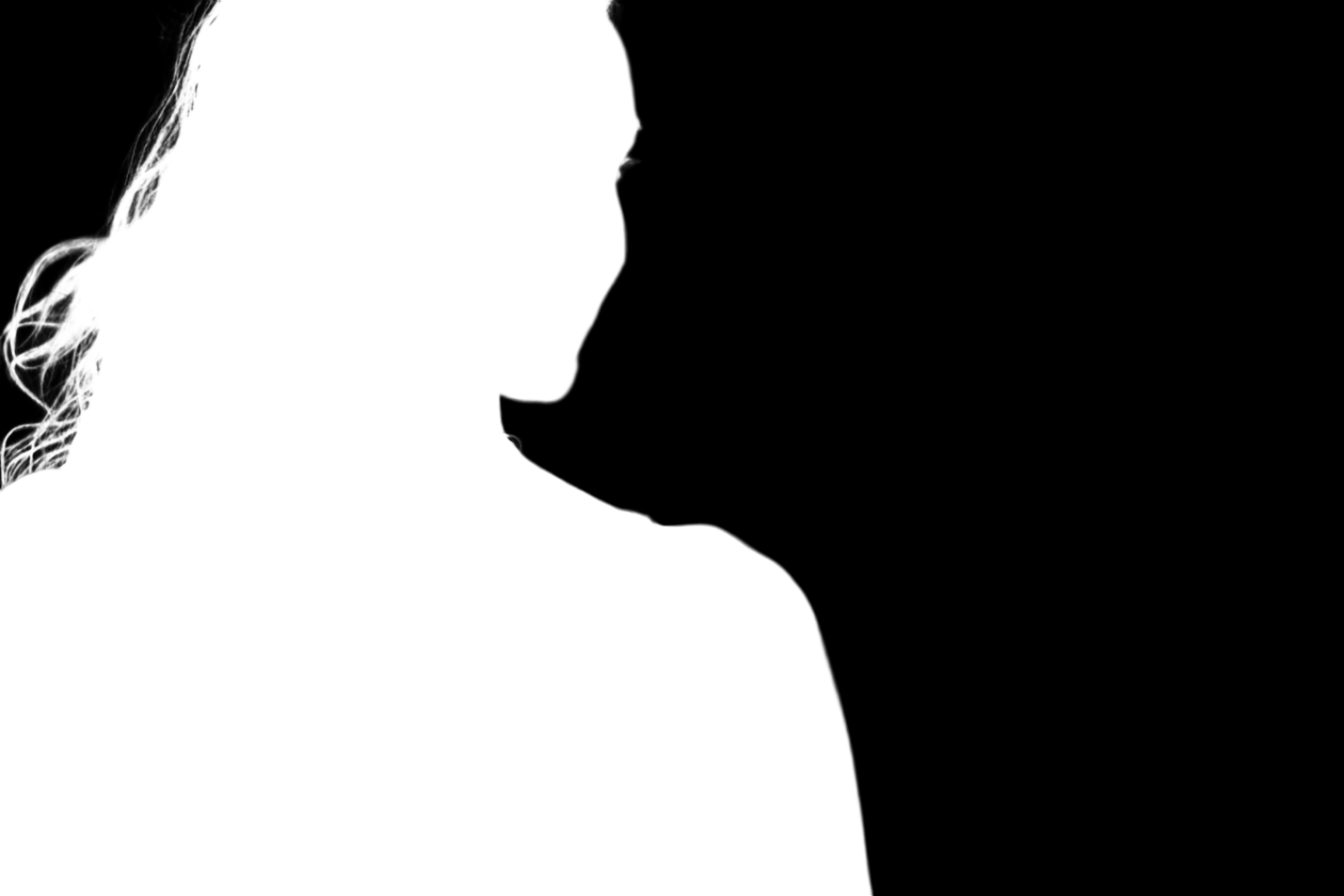}}
    \caption{Our}
    \label{fig:g_alpha}
  \end{subfigure}
  \centering
  \begin{subfigure}{0.16\linewidth}
    \centerline{\includegraphics[width=0.95\textwidth]{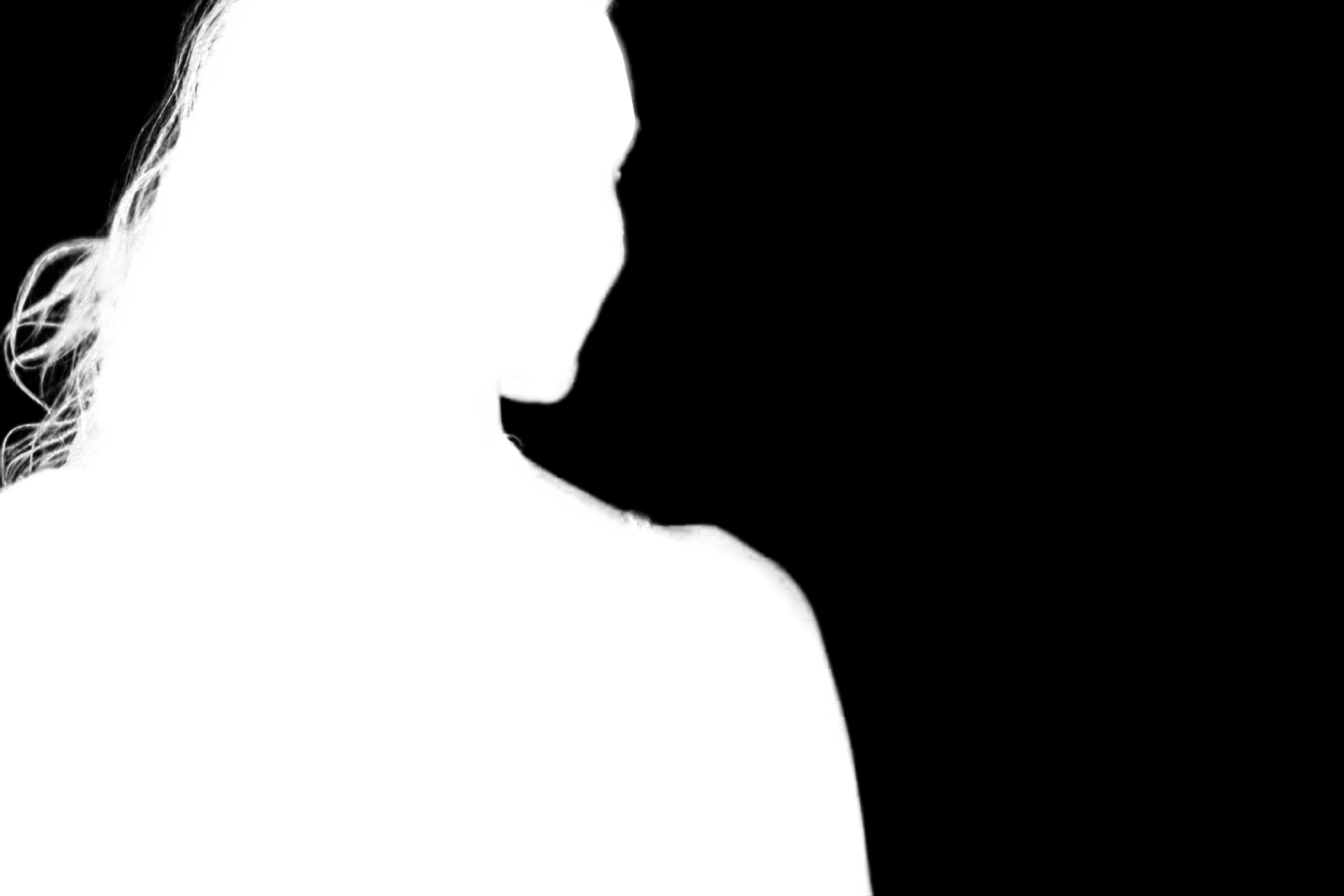}}
    \caption{GT}
    \label{fig:g_gt}
  \end{subfigure}
\centering
  \begin{subfigure}{0.16\linewidth}
    \centerline{\includegraphics[width=0.95\textwidth]{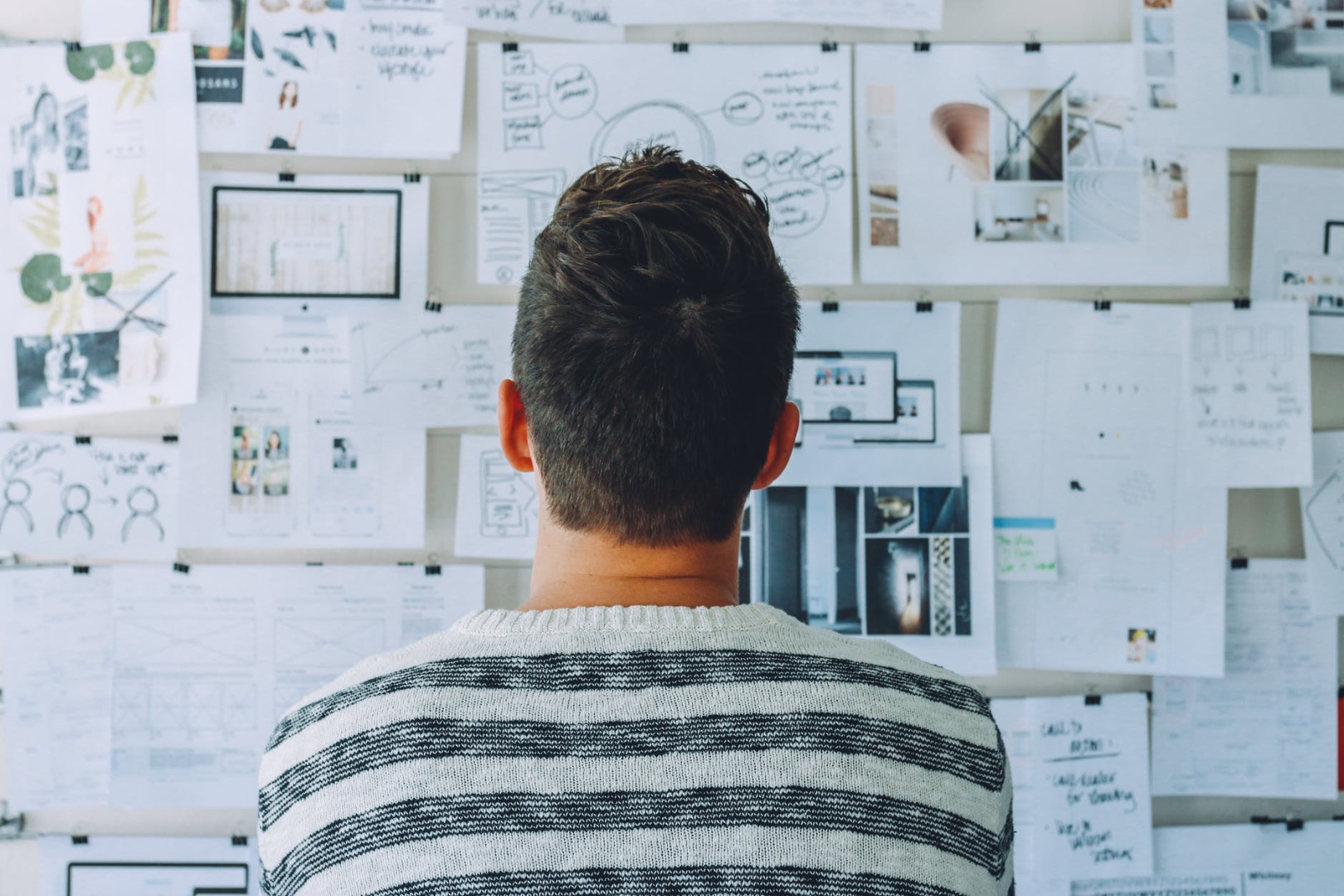}}
    \caption{Image}
    \label{fig:h_image}
  \end{subfigure}
  \centering
  \begin{subfigure}{0.16\linewidth}
    \centerline{\includegraphics[width=0.95\textwidth]{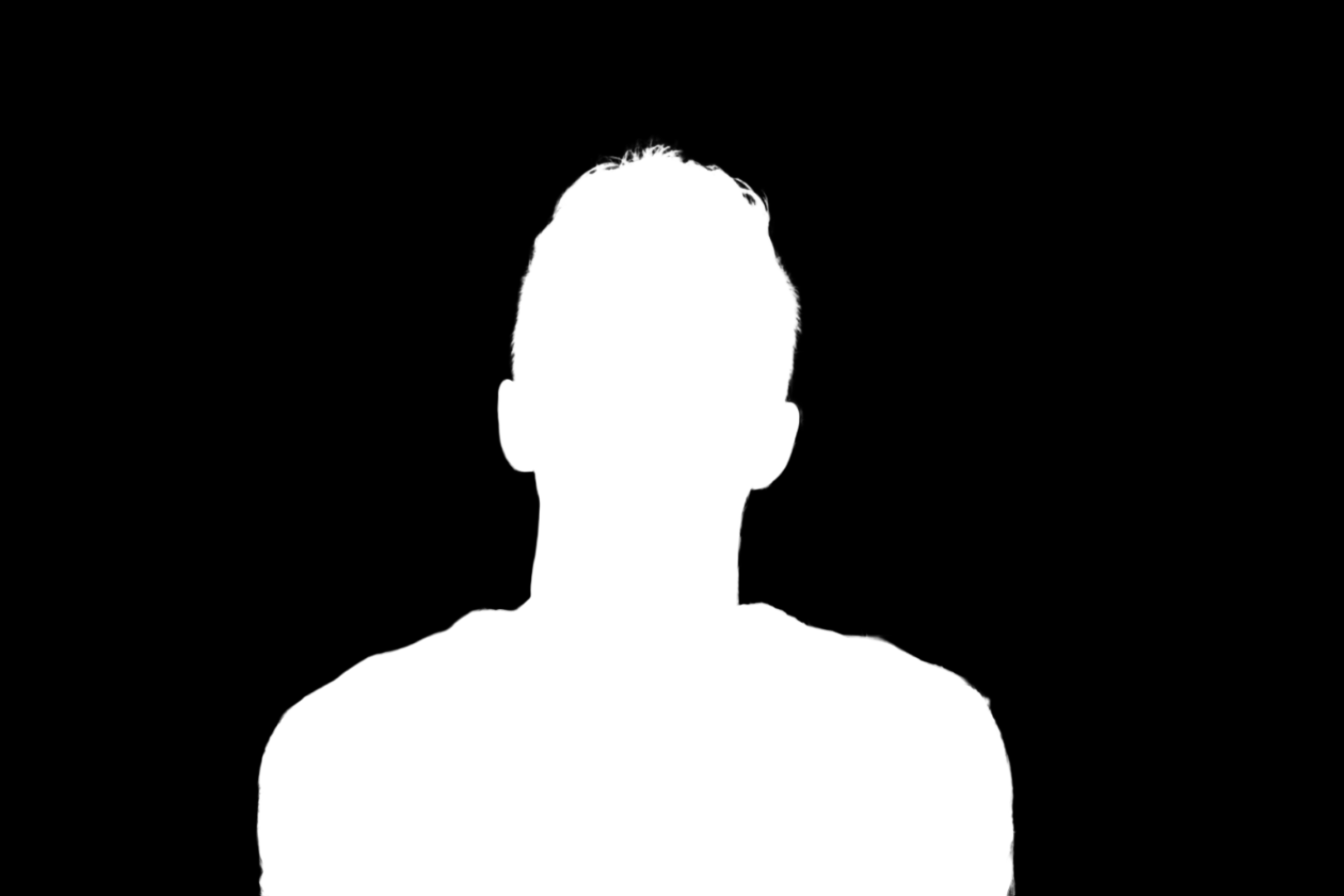}}
    \caption{Our}
    \label{fig:h_alpha}
  \end{subfigure}
  \centering
  \begin{subfigure}{0.16\linewidth}
    \centerline{\includegraphics[width=0.95\textwidth]{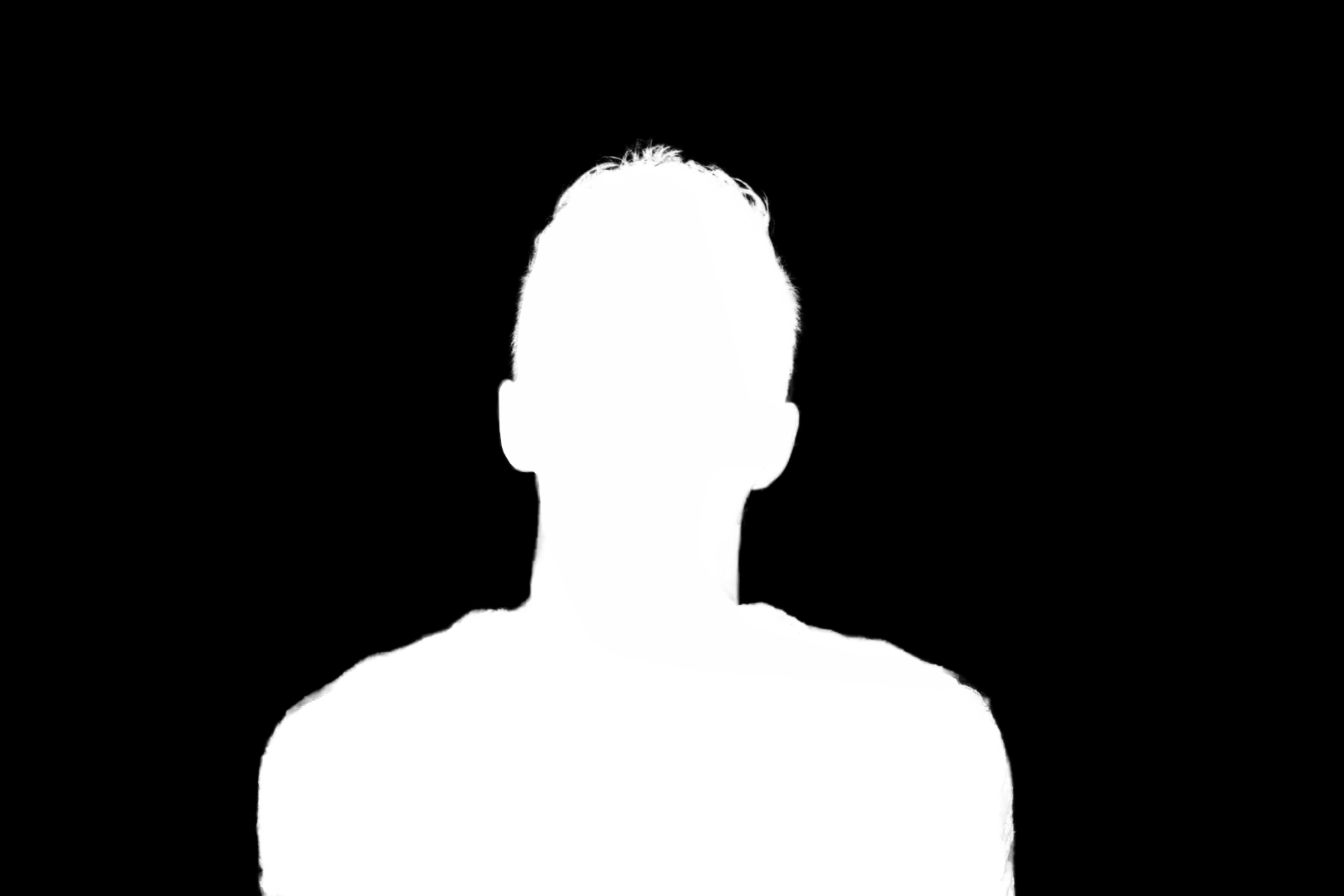}}
    \caption{GT}
    \label{fig:h_gt}
  \end{subfigure}
 
  \caption{Example of the real human image. The real image is from AIM-500\cite{li2021deep}.}
  \label{fig:human_example}
\end{figure*}

In natural image matting, human matting is a most popular application which is widely used in image composition, background replacement, and special effects. To evaluate the robustness of our PP-Matting in the real-world application, we collect and annotate more than 10k human images and collect 77k background images to train our model. Then we have a test in the real human images from AIM-500~\cite{li2021deep}. As shown in figure~\ref{fig:human_example}, the alpha mattes prediction results are comparable with the manual ground truth, which demonstrates our outstanding performance in the practical application.

\section{Conclusion}
In this work, we propose a trimap-free matting architecture named PP-Matting that can generate high-accuracy alpha matte with a single RGB image. Our method applies the high-resolution detail branch that extracts fine-grained details of the foreground. With the proposed semantic context branch (SCB), PP-Matting improves the detail prediction because of better semantic correctness. The proper interaction of the two branches using guidance flow helps the network achieve better semantic-aware prediction of alpha mattes. Extensive experiments on popular benchmarks and real-world datasets demonstrate our state-of-the-art performance.

{\small
\bibliographystyle{ieee_fullname}
\bibliography{egbib}
}

\end{document}